\documentclass[10pt,twocolumn,letterpaper]{article}

\usepackage[pagenumbers]{cvpr} 

\usepackage{graphicx}
\usepackage{amsmath}
\usepackage{amssymb}
\usepackage{booktabs}

\usepackage{multirow}
\usepackage{adjustbox}
\usepackage{algorithm}
\usepackage{algorithmic}
\usepackage{xcolor}
\usepackage{colortbl}
\usepackage[accsupp]{axessibility}
\usepackage{stackengine} 

\makeatletter
\usepackage{tikz}
\newcommand*\circled[2][1.6]{\tikz[baseline=(char.base)]{
    \node[shape=circle, draw, inner sep=0.8pt, 
        minimum height={1pt},] (char) {\vphantom{WAH1g}#2};}}
\makeatother

\definecolor{cvprblue}{rgb}{0.21,0.49,0.74}
\usepackage[pagebackref,breaklinks,colorlinks,allcolors=cvprblue]{hyperref}

\def\paperID{2} 
\def\confName{CVPR}
\def\confYear{2025}

\title{Improving Optical Flow and Stereo Depth Estimation \\ by Leveraging Uncertainty-Based Learning Difficulties}

\author{
Jisoo Jeong~~~
Hong Cai~~~
Jamie Menjay Lin~~~
Fatih Porikli~~~
\smallskip
\\
Qualcomm AI Research$^{\dagger}$~~~
\\
\smallskip
{\tt\small\{jisojeon, hongcai, jmlin, fporikli\}@qti.qualcomm.com \vspace{-12pt}}
}

\begin{document}
\maketitle
\begin{abstract} 

   Conventional training for optical flow and stereo depth models typically employs a uniform loss function across all pixels. However, this one-size-fits-all approach often overlooks the significant variations in learning difficulty among individual pixels and contextual regions. This paper investigates the uncertainty-based confidence maps which capture these spatially varying learning difficulties and introduces tailored solutions to address them. We first present the Difficulty Balancing (DB) loss, which utilizes an error-based confidence measure to encourage the network to focus more on challenging pixels and regions. Moreover, we identify that some difficult pixels and regions are affected by occlusions, resulting from the inherently ill-posed matching problem in the absence of real correspondences. 
   To address this, we propose the Occlusion Avoiding (OA) loss, designed to guide the network into cycle consistency-based confident regions, where feature matching is more reliable. By combining the DB and OA losses, we effectively manage various types of challenging pixels and regions during training. Experiments on both optical flow and stereo depth tasks consistently demonstrate significant performance improvements when applying our proposed combination of the DB and OA losses.
   
\end{abstract}

\section{Introduction}
\label{sec:intro}

{\let\thefootnote\relax\footnotetext{{
\hspace{-6.5mm} $\dagger$ Qualcomm AI Research is an initiative of Qualcomm Technologies, Inc.}}}

Feature matching serves as a core technique in the realm of computer vision, supporting various tasks and applications. Optical flow, which captures 2D pixel-wise displacements via feature matching, enables applications ranging from object tracking~\cite{kale2015moving}, action recognition~\cite{lee2018motion, cai2019temporal}, video compression~\cite{wu2018video, lu2019dvc}, and video frame interpolation~\cite{kong2022ifrnet, li2023amt, jeong2024ocai}. Similarly, rectified stereo depth estimation, discerning disparities between stereo images through feature matching, supports extensive applications including autonomous driving, extended reality, and mixed reality.

\begin{figure}[t]
\centering
\includegraphics[width=0.98\linewidth]{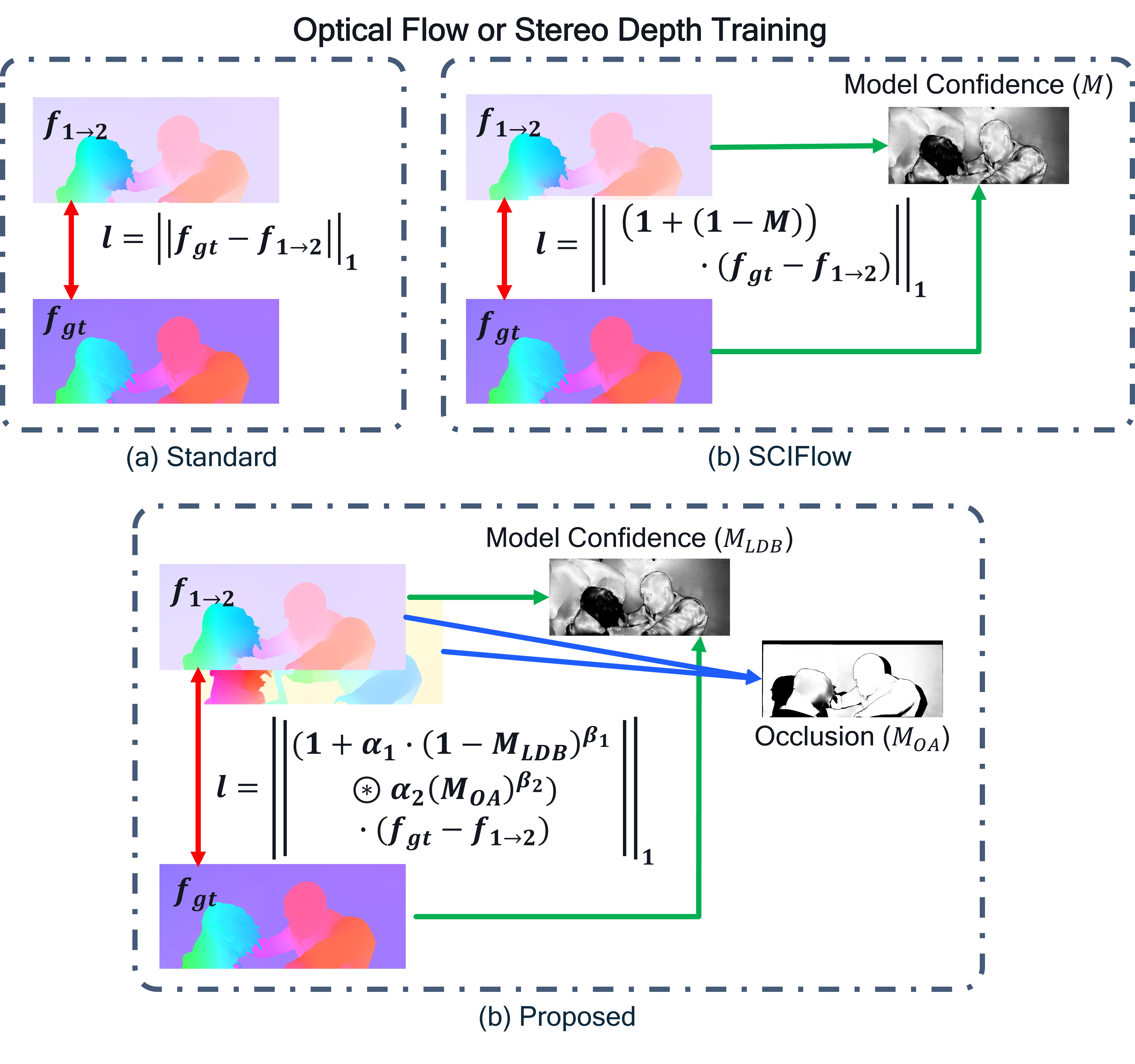}
\vspace{-3mm}
\caption{(a) Most existing methods (e.g.,~\cite{teed2020raft, lipson2021raft}) predominately treat training loss on each pixel equally for optical flow and Stereo depth. (b) SCIFlow~\cite{lin2024sciflow} utilizes a Regression Focal Loss, which focuses more on low-confident samples, for training optical flow models. (c) Our proposed approach more comprehensively considers two sources of learning difficulties in training, i.e., model confidence and occlusion.}
\vspace{-4mm}
\label{fig:intro}
\end{figure}

When training a network to predict optical flow~\cite{teed2020raft, jiang2021learning, huang2022flowformer} or stereo depth~\cite{lipson2021raft, li2022practical, karaev2023dynamicstereo}, it has been a standard practice in various models to use the same loss function across all valid pixels (Fig~\ref{fig:intro} top). In our study, however, the difficulty for the model to learn fine-grained correspondence could vary across pixels due to various factors including contexts,
non-rigid motions, and lighting conditions. Moreover, learning for occlusion regions can involve additional challenges~\cite{zhao2020maskflownet}.

In this paper, we propose a novel, effective training approach for optical flow and stereo depth models, leveraging uncertainty understanding to infer the learning difficulties for the model due to both pixel-wise contents and occlusions. In particular, we enable the model to learn by its confidence and to leverage contextual insight to mitigate the challenge of non-matching pixels.

More specifically, we first utilize a Difficulty Balancing (DB) loss, which imposes larger weights on pixels of lower prediction confidence. Our DB proposal is an extension from the Regression Focal Loss (RFL) originally introduced in~\cite{lin2024sciflow} for optical flow estimation. In this paper, our DB loss further improves over the RFL with optimal hyperparameters and we further extend the application of DB to stereo depth (Fig~\ref{fig:intro} bottom). 

Moreover, we introduce the Occlusion Avoiding (OA) loss, which mitigates the loss in regions where pixel-wise feature matching may not be feasible. In contrast to some prior works~\cite{jonschkowski2020matters, stone2021smurf} that completely discard non-matching areas during training, we continue to compute a minimum loss for such occluded areas, as it is still necessary to predict the motion for occlusion regions (from all object) in the dense prediction. In this paper, we utilize forward and backward consistency to derive occlusion information and infer matchable areas, and make the network concentrate its learning in the matchable regions during training.


Finally, we combine the DB and OA losses to address pixel-wise learning difficulties in training. Our proposal is model agnostic in nature. We apply our methods to several leading networks in optical flow and stereo depth, RAFT~\cite{teed2020raft}, FlowFormer~\cite{huang2022flowformer}, and RAFT-Stereo~\cite{lipson2021raft}. We empirically validate the effects of the DB and OA losses, and demonstrate how their proper combination yields further benefits in model accuracy.


In summary, our main contributions are as follows:

\begin{itemize}
\item We observe pixel-wise variations in learning difficulties and hypothesize the sources for such behavior in optical flow and stereo depth estimation.
\item First, we introduce the Difficulty Balancing (DB) loss to incorporate model confidence, which is inspired by and improves upon SCIFlow~\cite{lin2024sciflow}. We find the optimal hyperparameters for confidence map calibration, as well as weighting for optical flow and stereo estimation tasks.
\item We further propose the Occlusion Avoiding (OA) loss, which infers the matching reliability from the cycle (forward-backward) consistency and mitigates the weights of the regions that are less likely to be matchable accordingly. 
\item Finally, we demonstrate options to use both DB and OA losses simultaneously with an optimal combination, which provides uncertainty awareness in training and leads to improved model accuracy for optical flow and stereo depth. 


\end{itemize}


\section{Related Work}
\label{sec:related}

In optical flow estimation, consider two consecutive video frames, $I_{0}$ and $I_{1}$. We denote the optical flow from $I_{0}$ to $I_{1}$ as $f_{0 \rightarrow 1}$. 
In (rectified) stereo depth estimation, we consider two stereo images,$I_{L}$ and $I_{R}$. We denote the stereo depth estimation output as $d_{L \rightarrow R}$

\subsubsection{Optical Flow}
The architecture of RAFT \cite{teed2020raft} showed remarkable performance, and many subsequent studies \cite{huang2022flowformer, zhang2021separable, jiang2021learning} followed this baseline RAFT architecture. They extract features from two images and build a 4D correlation volume. And then, they iteratively regress to predict the optical flow output using ConvGRU blocks with the correlation volume. Such iteration-based prediction methods compute the loss for each optical flow prediction.

\begin{equation} \label{each_loss}
l^{i} = || (f_{gt} - f_{1 \rightarrow 2}^{i})||_{1}
\end{equation}

where $f_{gt}$ and $f_{1 \rightarrow 2} ^{i}$ are the optical flow ground truth and prediction for the \textit{i}-th iteration, respectively. The loss is accumulated over iterations as follows. 

\begin{equation} \label{total_loss}
L_{total} = \sum_{i=1}^{N} \gamma^{N-i} \cdot l^{i}
\end{equation}

\textbf{Stereo Depth Estimation: }
RAFT-stereo \cite{lipson2021raft} follows the RAFT architecture and achievs competitive quality in stereo depth estimation. It extracts features for the stereo images, builds 3D correlation volume, and iteratively updates the disparity (namely, the scaled inverse depth). The model is trained similarly to Eq.~\ref{each_loss} and \ref{total_loss} with the disparity predictions over iterations against the ground truth.

\begin{figure*}[ht]
\centering
\includegraphics[width=0.95\linewidth]{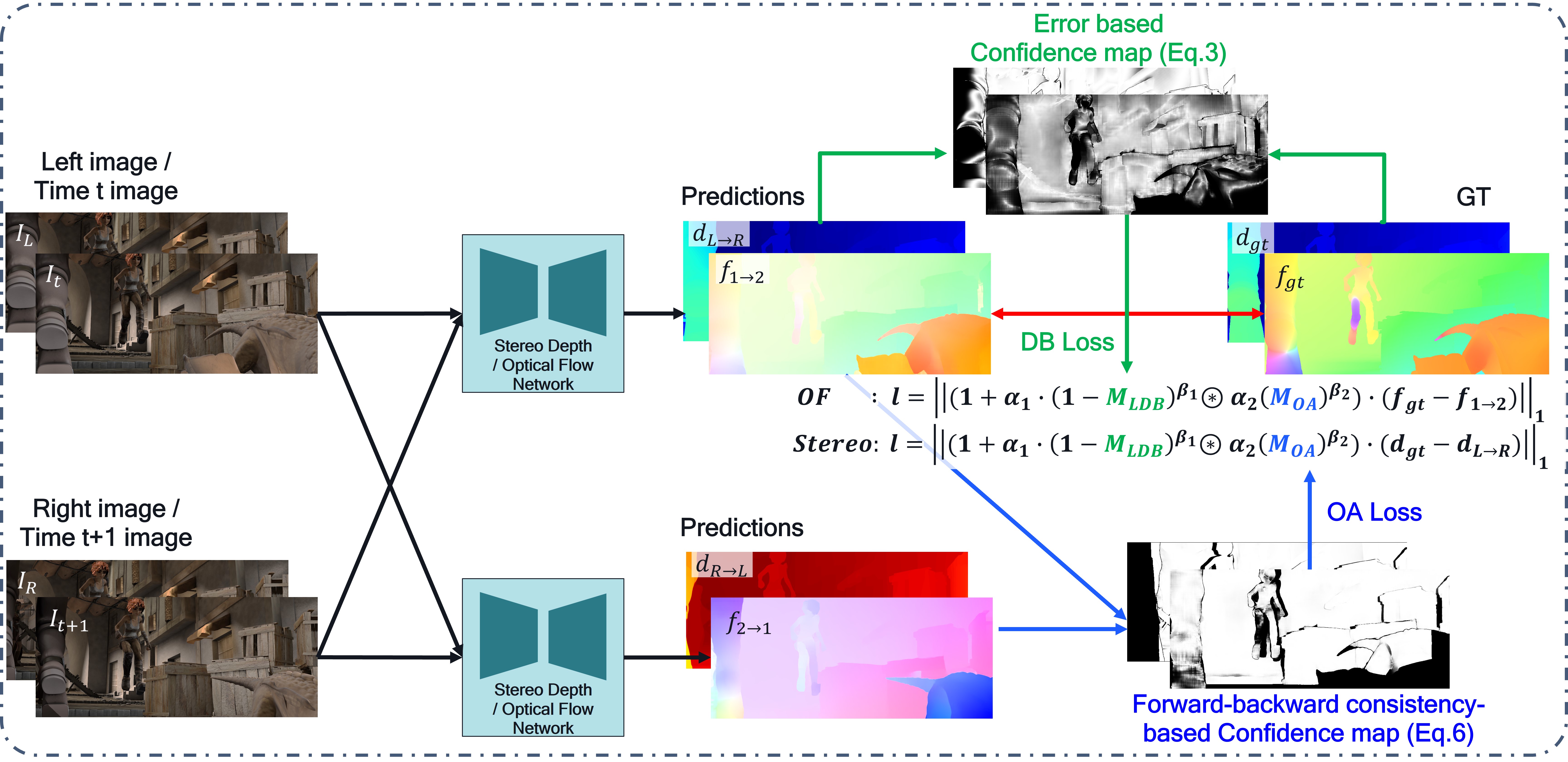}
\caption{Overview of our method. Optical Flows ($f_{1 \rightarrow 2}$ and $f_{2 \rightarrow 1}$) or Disparity ($d_{L \rightarrow R}$ and $d_{R \rightarrow L}$) are computed by the same model for the consecutive or stereo image pair. \textbf{Error map based Confidence map} is obtained using prediction and ground truth (Eq.~\ref{ldb_confmap}). \textbf{Forward backward consistency based Confidence map} is computed by Eq.~\ref{oa_confmap}. These confidence maps are used in the training loss. 
\circled[1.2]{$\ast$} represents the combination of two losses.}
\label{fig:method}
\end{figure*}

\textbf{Error-Based Confidence: }
LiteFlowNetV3 \cite{hui2020liteflownet3} proposed an error based confidence map as follows. 
\begin{equation} \label{ldb_confmap}
M(x)\! =\! \text{exp}{\big{(}-||f_{gt}(x) - f_{1 \rightarrow 2}(x)||^2\big{)}} 
\vspace{-2pt}
\end{equation}
In contrast to LiteFlowNetV3, which incorporated a confidence map into its architecture, SCIFlow\cite{lin2024sciflow} employs an error-based confidence map to derive a weighted loss in training.

\begin{equation} \label{ldb_loss}
l_{rfl}^{i} = || (1 + (1 - M)) \cdot (f_{gt} - f_{1 \rightarrow 2}^{i})||_{1}
\vspace{-2pt}
\end{equation}
In this paper, we take the RFL loss as our baseline loss to further search for the optimal form for the weighting to derive a confidence map. We also extend our definition of the DB loss to stereo depth for benefits in model accuracy.

\textbf{Cycle Consistency-Based Confidence: }
DistractFlow~\cite{jeong2023distractflow} proposed a cycle consistency-based confidence map using forward-backward consistency check~\cite{meister2018unflow}. The consistency check~\ref{occlusion_mask} and the confidence map~\ref{oa_confmap} are computed as follows. 
\begin{equation} \label{occlusion_mask}
\begin{split}
|\widehat{f}_{1 \rightarrow 2}(x)\! +\! \widehat{f}_{2 \rightarrow 1}(x\!+\!\widehat{f}_{1 \rightarrow 2}(x))|^2 \qquad \qquad \qquad\\
\qquad < \gamma_{1} (|\widehat{f}_{1 \rightarrow 2}|^2\! +\! |\widehat{f}_{2 \rightarrow 1}(x\!+\!\widehat{f}_{1 \rightarrow 2})|^2 )\! + \!\gamma_{2}
\end{split}
\vspace{-2pt}
\end{equation}
\begin{equation} \label{oa_confmap}
M(x)\! =\! \text{exp}  {\Bigg{(}\!\!\!-\!\frac{|\widehat{f}_{1 \rightarrow 2}(x)\! +\! \widehat{f}_{2 \rightarrow 1}(x\!+\!\widehat{f}_{1 \rightarrow 2}(x))|^2}{\gamma_{1} (|\widehat{f}_{1 \rightarrow 2}|^2\! +\! |\widehat{f}_{2 \rightarrow 1}(x\!+\!\widehat{f}_{1 \rightarrow 2})|^2 )\! + \!\gamma_{2} } \Bigg{)}},
\vspace{-2pt}
\end{equation}
where $\gamma_1$ and $\gamma_2$ are set to 0.01 and 0.5, respectively. Fixmatch~\cite{sohn2020fixmatch} style pseudo-label based semi-supervised optical flow methods~\cite{jeong2023distractflow, jeong2024ocai} created pseudo labels based on the confidence map to train the model.

A confidence map (Eq.~\ref{oa_confmap}) may be derived based on a cycle (namely, through forward warping and then backward warping) consistency check. The concept is that, a region is likely non-occluded if its features is cycle consistent and vise versa

\section{Method}
\label{sec:method}
We identify two main sources of learning difficulty during optical flow and stereo depth model training. First, due to various factors such as non-rigid motions, brightness changes, and large motions, it can be more difficult for the model to learn and predict for such regions. It may help to encourage the network to focus more on fitting these hard samples, e.g., by placing more weights on the corresponding losses. Therefore, we introduce our Difficulty Balancing (DB) loss in Section~\ref{sec:DB}. 
There is, however, another type of difficulty due to occlusion (or lack of correspondences) in certain regions. Simply forcing the network to predict in such regions of ill conditions can cause overfitting and skew the network's learning.~\cite{zhao2020maskflownet} To address such challenge, we separately propose the Occlusion Avoiding (OA) loss (Section~\ref{sec:OA}) with proper insight to help the network learn to cope with challenges in such regions. Finally, we further combine the DB and OA losses to leverage benefits of both (Section~\ref{sec:DB+OA}).

\subsection{Difficulty Balancing (DB)}\label{sec:DB}

We propose the Difficulty Balancing (DB) loss, built on top of the RFL~\cite{lin2024sciflow} in Eq.~\ref{ldb_loss} for optical flow networks, and extend it to both optical flow and stereo depth estimation to balance the learning difficulty, where $M_{db}$ is computed using Eq.~\ref{ldb_confmap}:

\begin{equation} \label{ldb_dfs_loss}
\footnotesize
\begin{split}
OF \quad \ : \quad l_{DB}^{i} = || (1 + \alpha \cdot (1 - M_{DB})^{\beta}) \cdot (f_{gt} - f_{1 \rightarrow 2}^{i})||_{1}, \ \ \\
Stereo : \quad l_{DB}^{i} = || (1 + \alpha \cdot (1 - M_{DB})^{\beta}) \cdot (d_{gt} - d_{L \rightarrow R}^{i})||_{1},
\end{split}
\end{equation} 

We set ($\alpha$,  $\beta$) to (2.0, 0.5) for optical flow, and (2.0, 1.0) for stereo depth estimation, respectively (see Tables~\ref{tab:ldb_ablation} and~\ref{tab:dfs_training_ablation} for our empirical studies). Intuitively, when the prediction is close to the ground truth, $M_{DB}$ can be close to 1. In such cases, the DB loss operates similarly to the standard L1 loss. On the other hand, when the prediction has a large difference from the ground truth, the confidence score $M_{db}$ can go as low as 0, which in effect places significantly larger weights in the training loss values for such regions. 

\subsection{Occlusion Avoiding (OA)}\label{sec:OA}
We further define the Occlusion Avoiding (OA) loss by leveraging the cycle (forward-backward) consistency for insight to guide the network regarding occlusions:  
\begin{equation} \label{oa_loss}
\footnotesize
\begin{split}
OF \quad \ : \quad l_{OA}^{i} = || (1 + \alpha \cdot (M_{OA})^{\beta}) \cdot (f_{gt} - f_{1\rightarrow 2}^{i})||_{1},  \\
Stereo : \quad l_{OA}^{i} = || (1 + \alpha \cdot (M_{OA})^{\beta}) \cdot (d_{gt} - d_{L \rightarrow R}^{i})||_{1},
\end{split}
\end{equation}
where $M_{OA}$ is computed by Eq.~\ref{oa_confmap}. 

We set ($\alpha$,  $\beta$) to (2.0, 1.0) for optical flow, and (1.0, 1.0) for stereo depth estimation, respectively; Tables~\ref{tab:oa_ablation} and~\ref{tab:dfs_training_ablation} provide our empirical studies. Intuitively, if the forward and backward flows collectively form a consistent cycle in terms of the feature consistency, such region can be weighted more in the computed loss for the region. On the other hand, if the cycle consistency is low, the loss for such region will be similar to the standard L1 loss. 

\begin{figure}[t]
\centering
\includegraphics[width=0.95\linewidth]{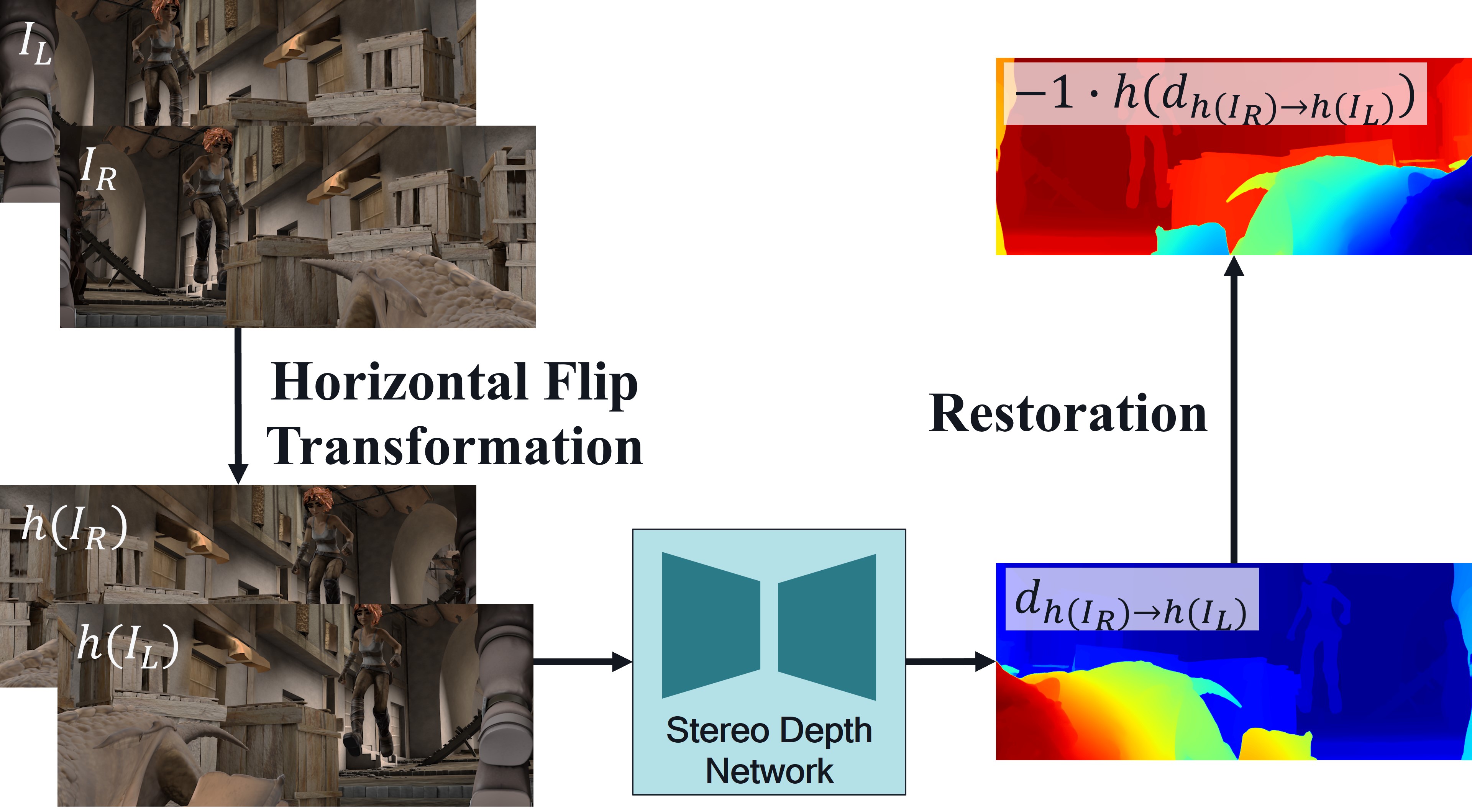}
\caption{The \textbf{Transformation-and-Restoration} technique to obtain the reverse (right-to-left) disparity. }
\label{fig:stereo_oa}
\end{figure}

In stereo depth, a confidence map is similarly derived by computing $M_{OA}$ using the left-to-right and right-to-left disparity maps for warping. One notable detail, however, is that optical flow is for 2D signed displacements, whereas stereo disparity is 1D and unsigned (typically from left to right). Therefore, $M_{OA}$ requires an additional pass of estimation for the reverse (namely, right-to-left) disparity. To address this requirement, we employ a transformation-and-restoration operation as shown in Fig~\ref{fig:stereo_oa}, where we swap and horizontal flip (\textit{h}) the $I_{L}$ and $I_{R}$ stereo images and estimate the disparity ($d_{h(I_{R}) \rightarrow h(I_{L})}$) between the flipped right image ($I_{R}$) and the flipped left image ($I_{L}$). And then, we restore the disparity value by flipping and changing its sign ($-1 \cdot h(d_{h(I_{R}) \rightarrow h(I_{L})}$)). This technique allows us to reuse our original stereo depth model to obtain the reverse disparity from $I_{R}$ to $I_{L}$ and generate an OA weighting map.

\begin{table*}[ht]
\begin{center}
\caption{Optical flow results on Sintel (train) and KITTI (train) datasets. We train the model on FlyingChairs (C) and FlyingThings3D (T).
\textbf{Bold}/\underline{Underline}: Best and second best results. (* is tested by ourselves, and $\dagger$ is obtained via the tile technique~\cite{huang2022flowformer}.)
}
\vspace{-2mm}
\label{tab:of_training}
\adjustbox{max width=0.9\textwidth}
{
\begin{tabular}{|l|l||c|c|c|c|}
\hline
\multirow{2}{*}{Model} & \multirow{2}{*}{Method} & \multicolumn{2}{|c|}{Sintel (train)} & \multicolumn{2}{|c|}{KITTI (train)}\\
\cline{3-6}
& & Clean-EPE ($\downarrow$) & Final-EPE ($\downarrow$) & EPE ($\downarrow$) & Fl-all ($\downarrow$) \\
\hline
\multirow{7}{*}{RAFT~\cite{teed2020raft}} & Baseline & 1.43 / 1.43* & 2.71 / 2.69* & 5.04 / 5.00*  & 17.4 / 17.45* \\
\cline{2-6}
& Difficulty Balancing (DB) & 1.41 & 2,68 & 4.65 & 15.92 \\
& Occlusion Avoiding (OA) & \textbf{1.34} & 2.66 & \textbf{4.44} & 15.77\\
\cline{2-6}
& Combination (Sum) & 1.39 & 2.70 & 4.72 & 16.55 \\
& \cellcolor{lightgray}Combination (Multiplication) & \cellcolor{lightgray}\underline{1.35} &  \cellcolor{lightgray}\underline{2.65} &  \cellcolor{lightgray}\underline{4.50} &  \cellcolor{lightgray}\textbf{15.45} \\
& Combination (Masking) & 1.37 & 2.70 & 4.59 & \underline{15.65}\\
& Combination (Mask-Sum) & 1.40 & \textbf{2.57} & 4.59 & 16.01\\
\hline
\multirow{2}{*}{FlowFormer~\cite{huang2022flowformer}} & Baseline & 1.01 / 0.98* & 2.40 / \textbf{2.34}* & 4.09$^{\dagger}$ / 4.26*$^{\dagger}$  & 14.72$^{\dagger}$ / 14.47*$^{\dagger}$ \\
& \cellcolor{lightgray}Combination (Multiplication) & \cellcolor{lightgray}\textbf{0.97} &  \cellcolor{lightgray}2.35 &  \cellcolor{lightgray}\textbf{4.03$^{\dagger}$} &  \cellcolor{lightgray}\textbf{14.17$^{\dagger}$} \\
\hline
\end{tabular}
}

\vspace{-3mm}
\end{center}
\end{table*}

\subsection{Combination of DB and OA}\label{sec:DB+OA}
In order to simultaneously benefit from both DB and OA losses, we explore multiple options to combine them.
We first directly combined DB and OA losses in an additive and multiplicative ways ($l_{sum}$ and $l_{mul}$). 
This naive approach, however, cancels out the effects of both losses after summation, as DB and OA tend to react oppositely given a (high or low) confidence score.
To address this issue, we apply a hard mask for the DB loss in the occluded region using $M_{OA}$ ($l_{mask}$), and then we sum the masked DB and OA losses ($l_{mask-sum}$).
\begin{equation} \label{sum_loss}
\footnotesize
\begin{split}
l_{sum}^{i} = \qquad \qquad \qquad \qquad \qquad \qquad \qquad \qquad \qquad \qquad \qquad \qquad \qquad \\
|| (1+ \alpha_{1} \cdot (1-M_{DB})^{\beta_{1}} + \alpha_{2} \cdot (M_{OA})^{\beta_{2}} )\cdot (f_{gt} - f_{1\rightarrow 2}^{i})||_{1} \\
l_{mul}^{i} = \qquad \qquad \qquad \qquad \qquad \qquad \qquad \qquad \qquad \qquad \qquad \qquad \qquad \\
|| (1+ \alpha_{1} \cdot (1-M_{DB})^{\beta_{1}} \cdot \alpha_{2} \cdot (M_{OA})^{\beta_{2}} )\cdot (f_{gt} - f_{1\rightarrow 2}^{i})||_{1} \\
l_{mask}^{i} = \qquad \qquad \qquad \qquad \qquad \qquad \qquad \qquad \qquad \qquad \qquad \qquad \quad \ \\
|| (1+ H(M_{OA}) \cdot \alpha_{1} \cdot (1-M_{DB})^{\beta_{1}}) \cdot (f_{gt} - f_{1\rightarrow 2}^{i})||_{1} \\
l_{mask-sum}^{i} = \qquad \qquad \qquad \qquad \qquad \qquad \qquad \qquad \qquad \qquad \qquad \quad \\
|| (1 + H(M_{OA}) \cdot \alpha_{1} \cdot (1-M_{DB})^{\beta_{1}} + \qquad \qquad \qquad \\ 
\qquad \qquad \alpha_{2} \cdot (M_{OA})^{\beta_{2}}) \cdot (f_{gt} - f_{1\rightarrow 2}^{i})||_{1}
\end{split}
\end{equation}
where, $H(M_{OA})$ is the hard masking operation for the occlusion obtained in Eq.~\ref{occlusion_mask}.
We then accumulate the loss from each iteration as in Eq.~\ref{total_loss}.

\begin{figure*}[t]
\begin{center}$
\centering
\begin{tabular}{cccc}
\vspace{-0.1cm}
\hspace{-0.3cm}
\rotatebox{90}{\quad \text{\small{\textbf{Baseline}}}} &
\hspace{-0.45cm} \includegraphics[width=5.75cm]{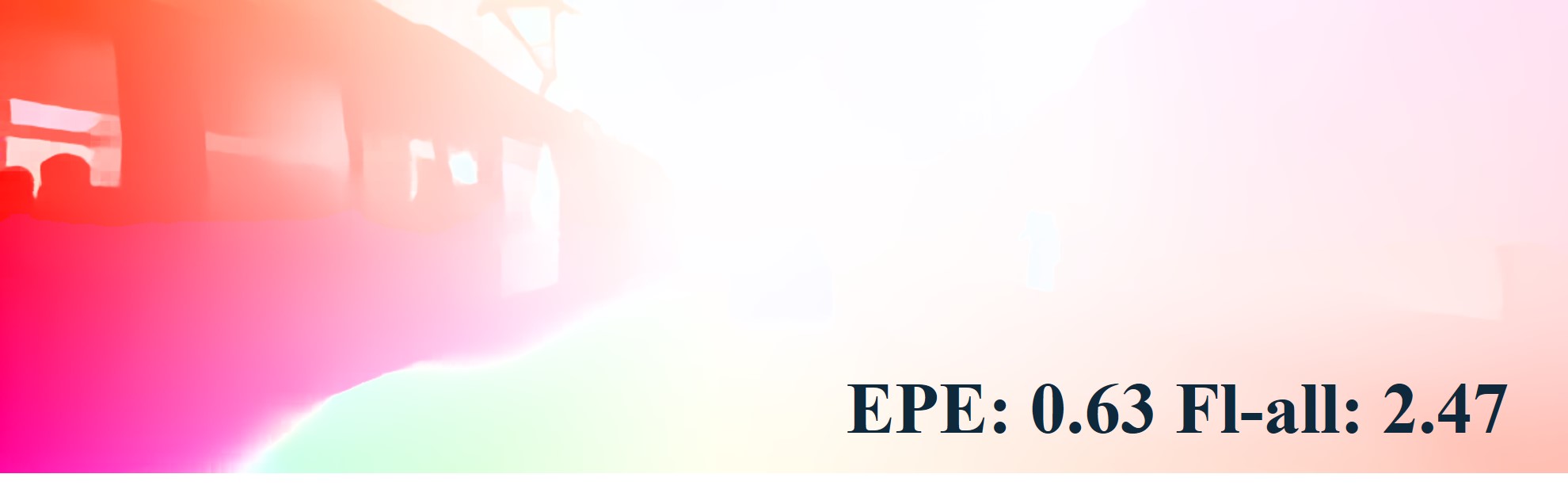}
& \hspace{-0.45cm} \includegraphics[width=5.75cm]{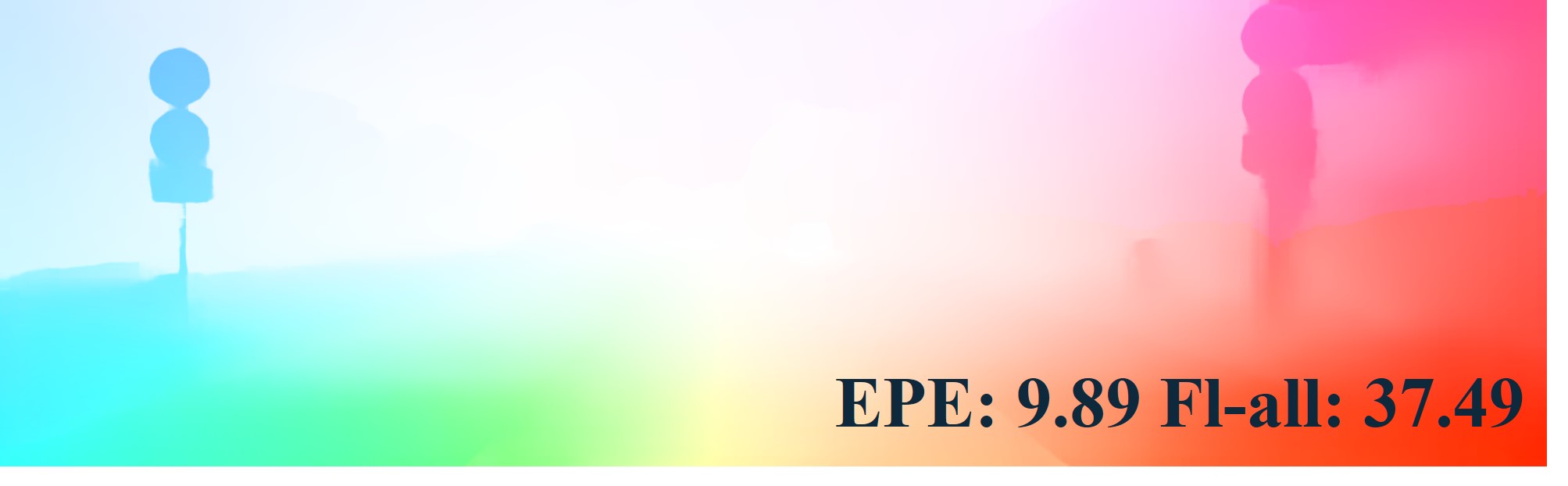}
& \hspace{-0.5cm} \includegraphics[width=5.75cm]{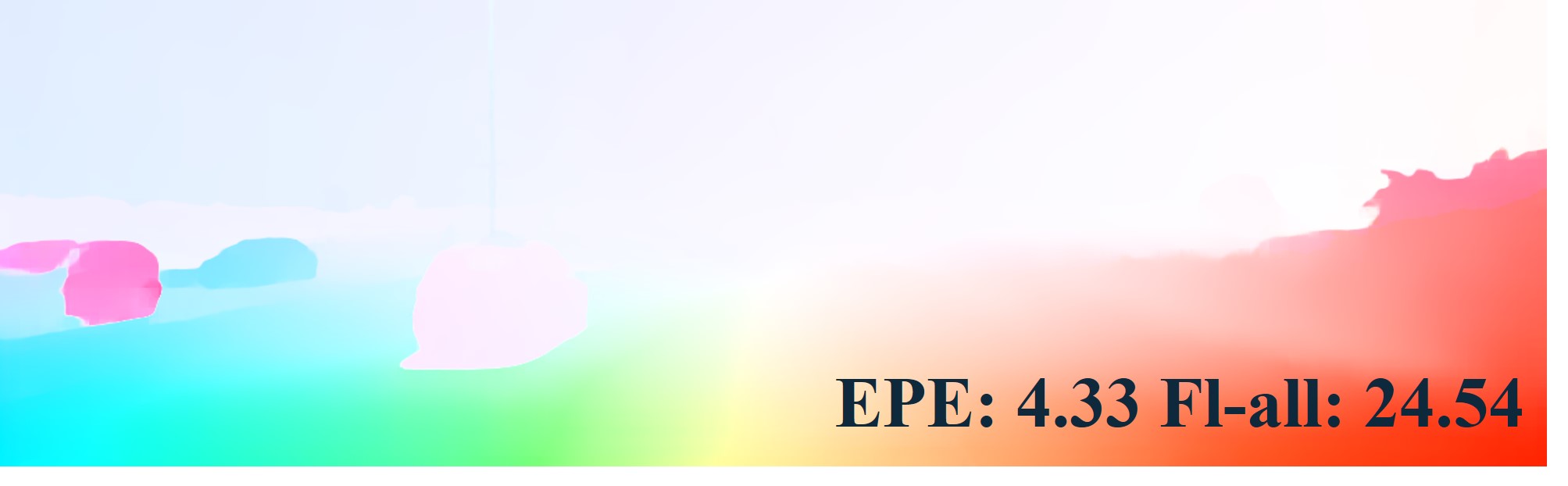} \\
\vspace{-0.1cm}
\hspace{-0.3cm} \rotatebox{90}{\qquad \ \text{\textbf{DB}}} &
\hspace{-0.45cm} \includegraphics[width=5.75cm]{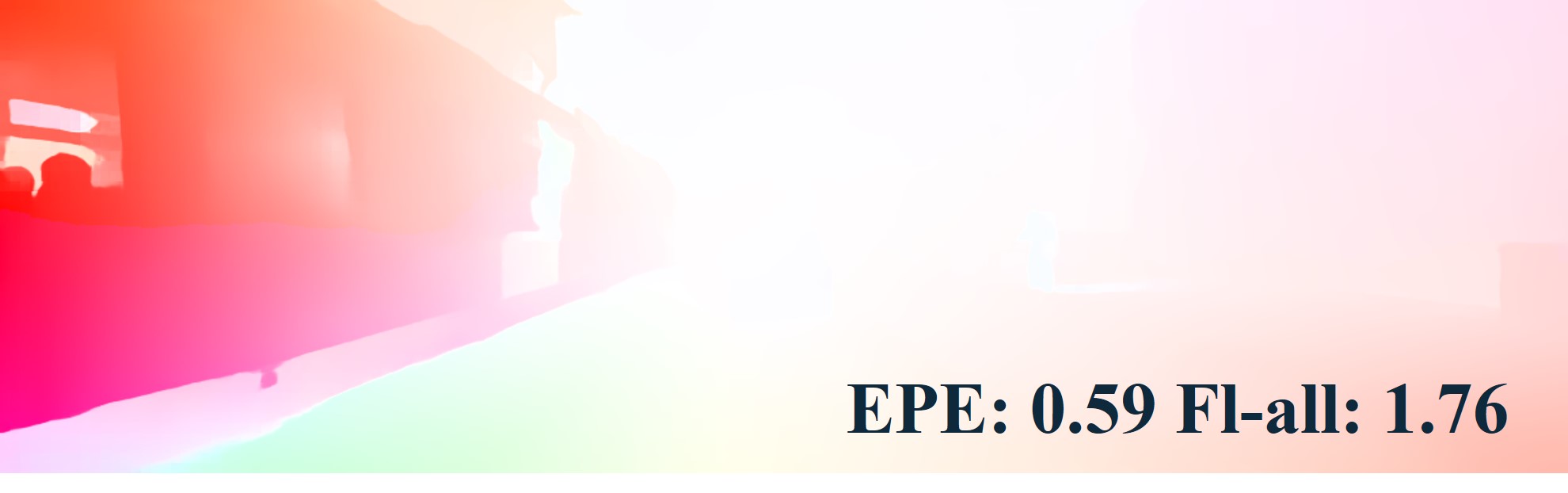}
& \hspace{-0.45cm} \includegraphics[width=5.75cm]{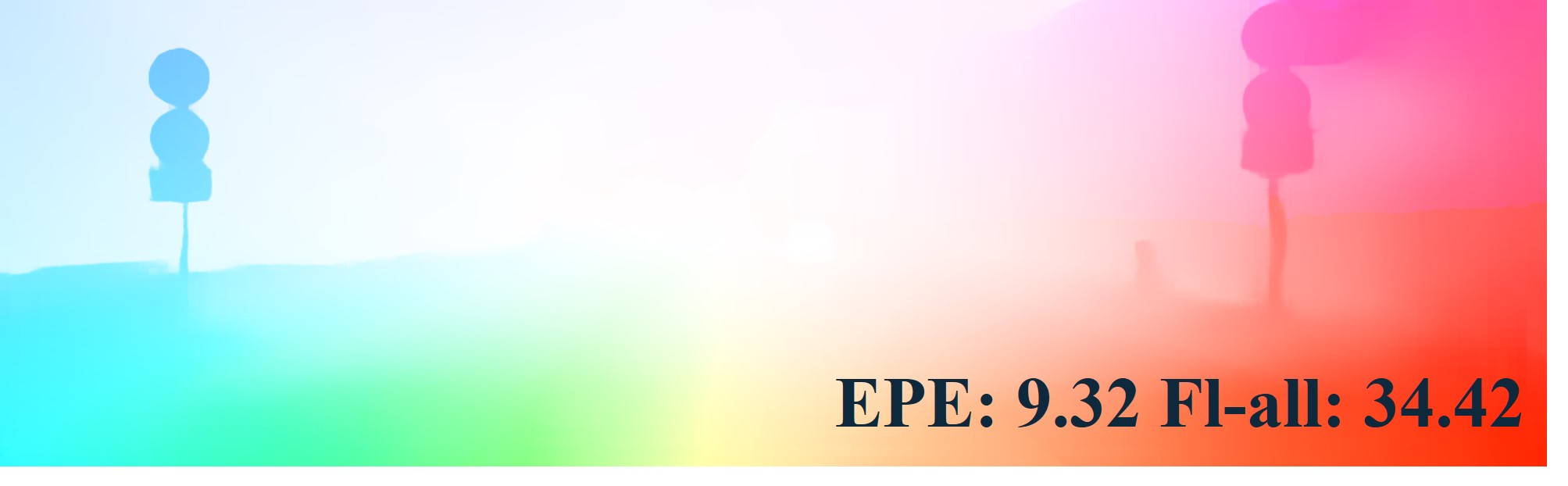}
& \hspace{-0.5cm} \includegraphics[width=5.75cm]{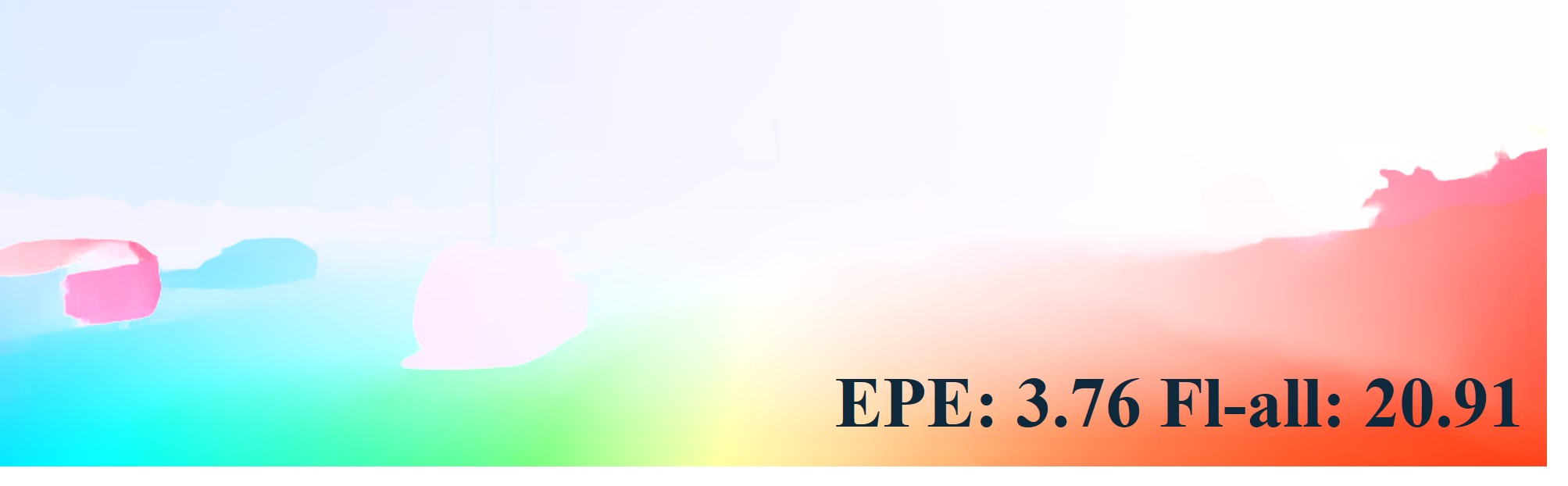} \\
\vspace{-0.1cm}
\hspace{-0.3cm}  \rotatebox{90}{\qquad \text{\textbf{OA}}} &
\hspace{-0.45cm} \includegraphics[width=5.75cm]{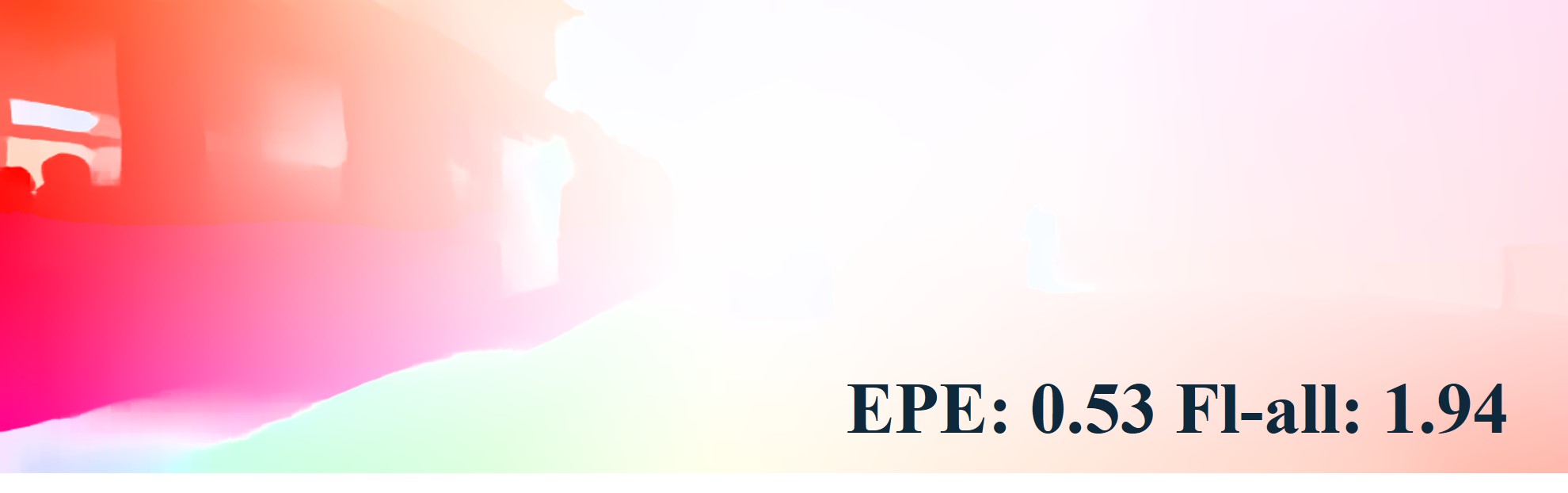}
& \hspace{-0.45cm} \includegraphics[width=5.75cm]{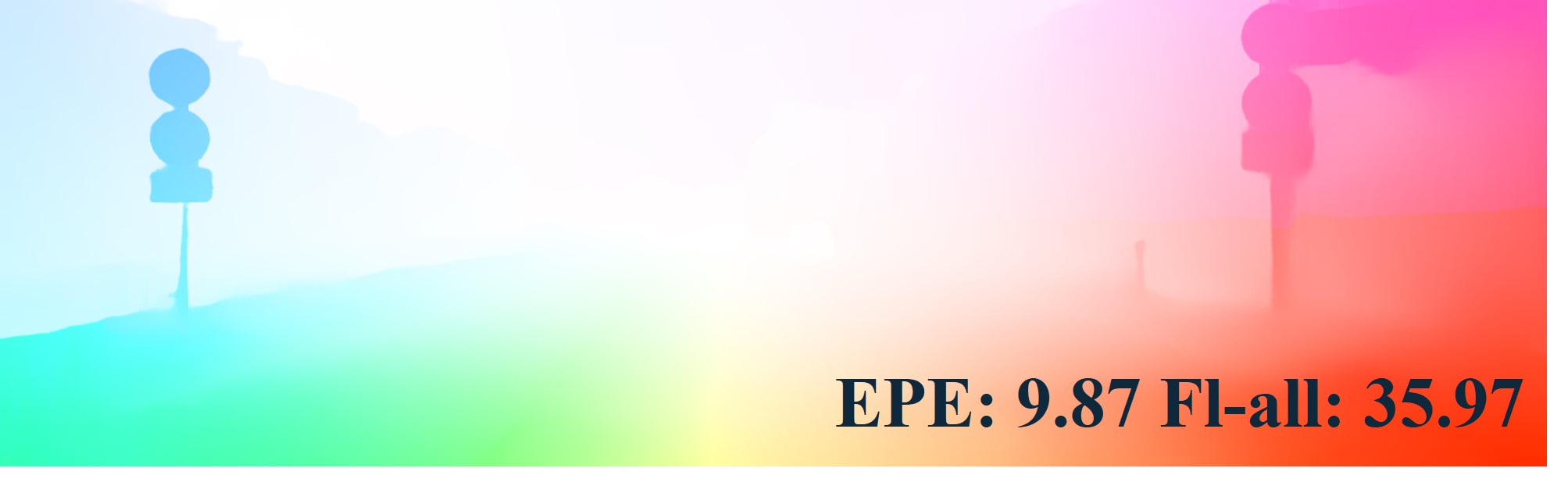}
& \hspace{-0.5cm} \includegraphics[width=5.75cm]{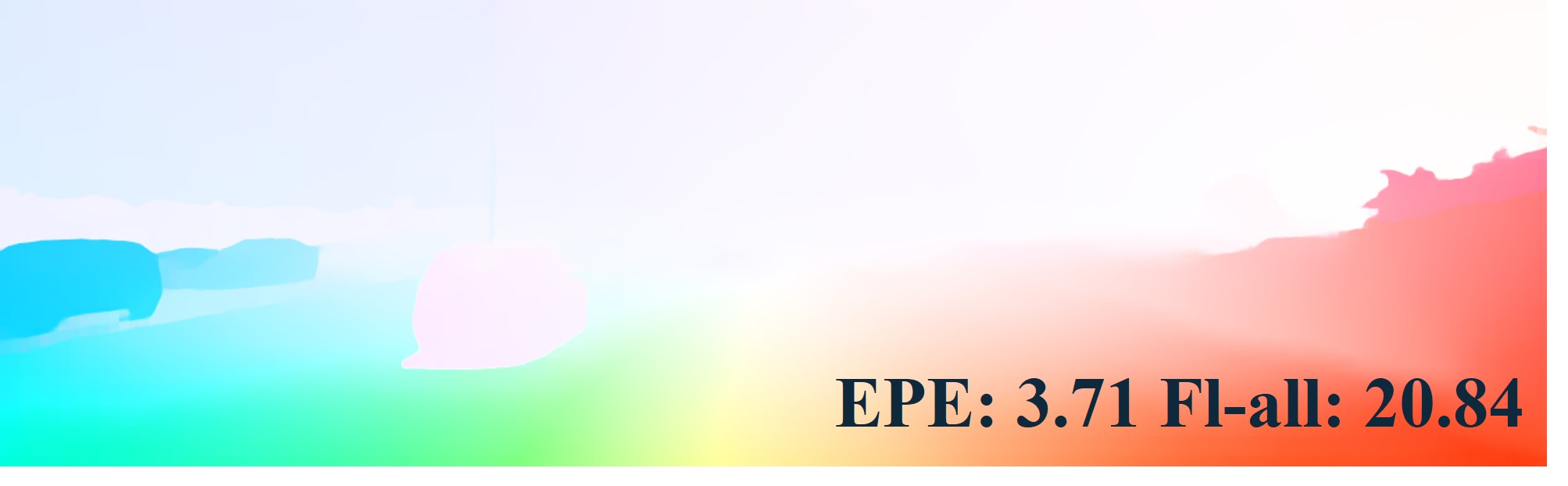} \\
\vspace{-0.1cm}
\hspace{-0.3cm} \rotatebox{90}{\quad \ \text{\textbf{Comb}}} &
\hspace{-0.45cm} \includegraphics[width=5.75cm]{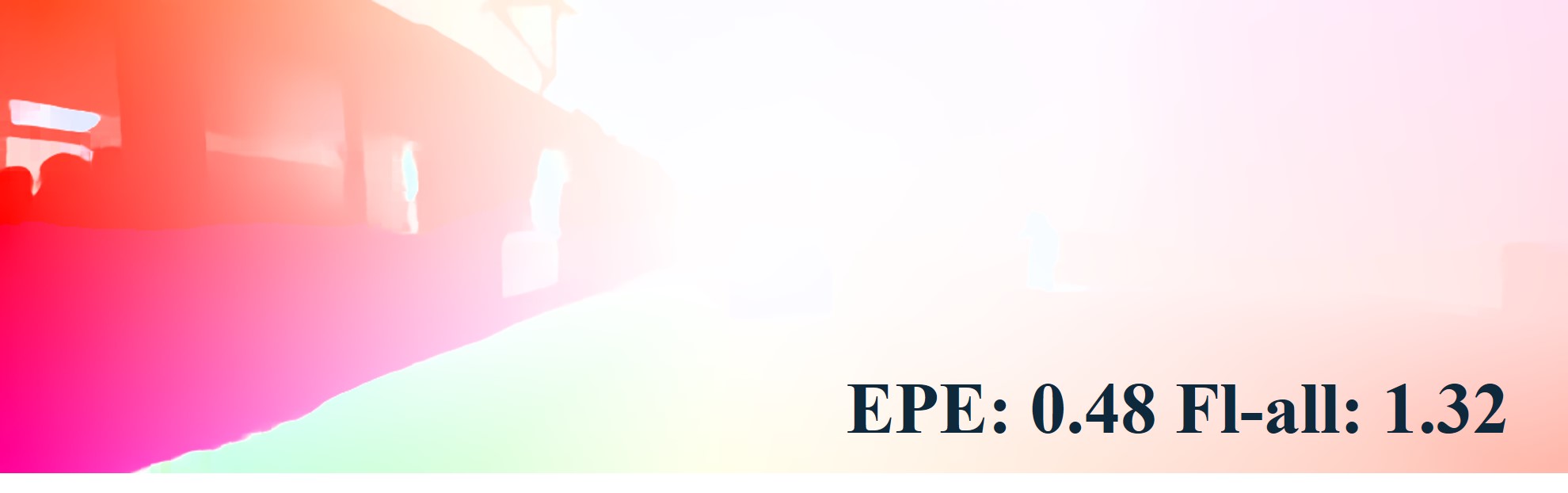}
& \hspace{-0.45cm} \includegraphics[width=5.75cm]{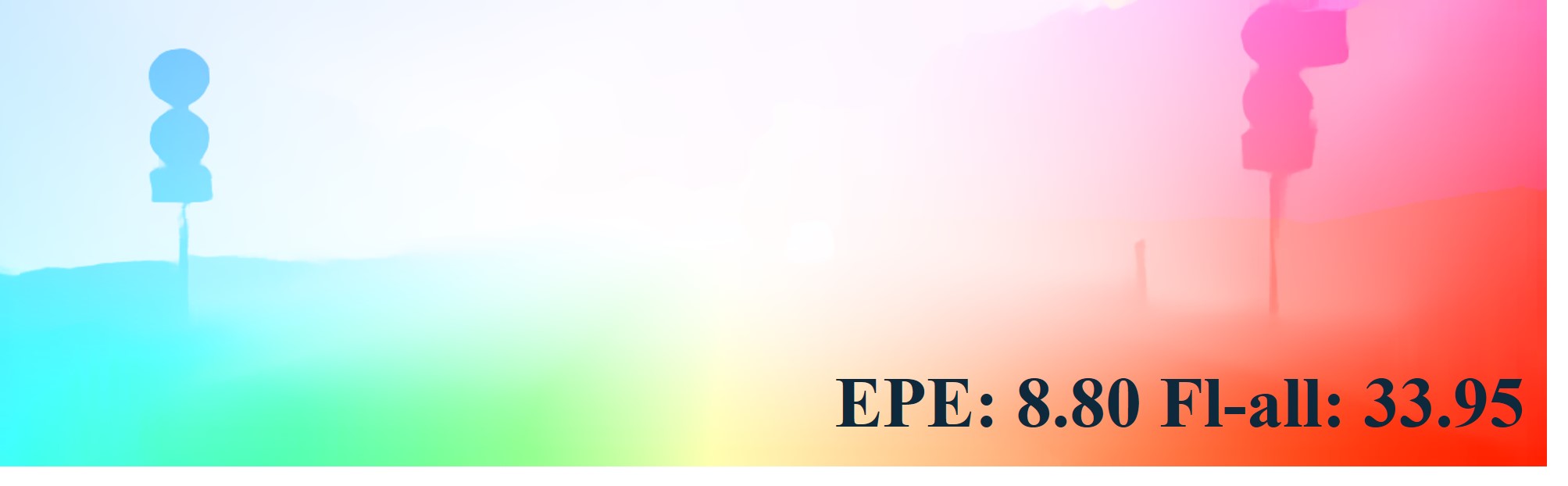}
& \hspace{-0.5cm} \includegraphics[width=5.75cm]{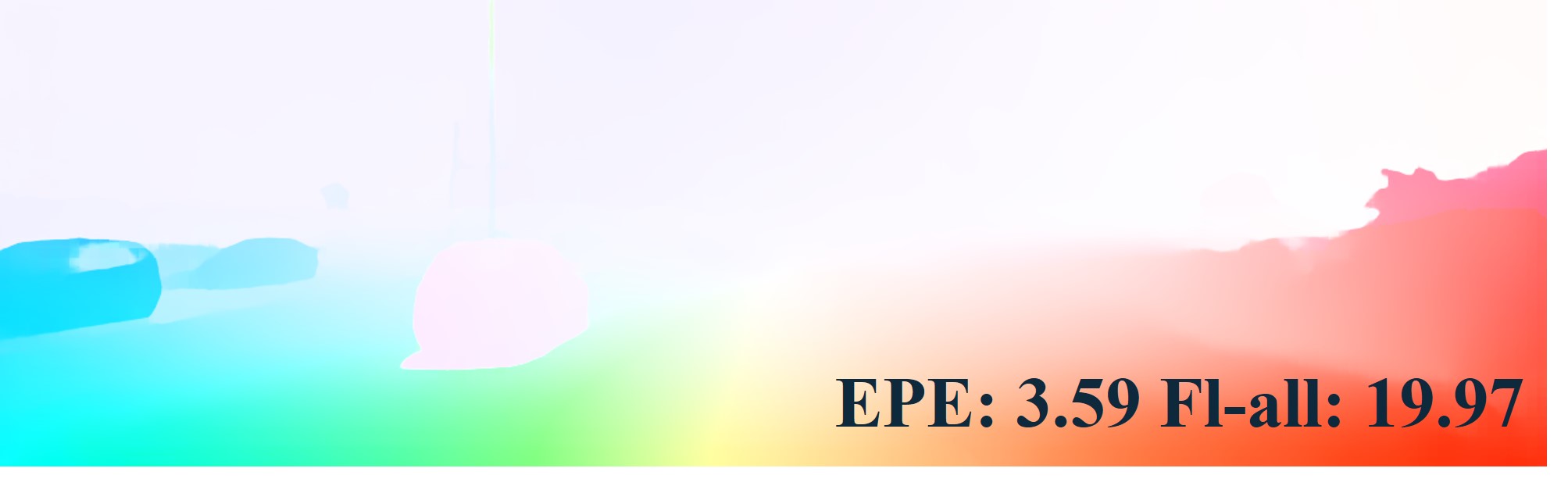} \\
\end{tabular}$
\end{center}
\vspace{-15pt}
\caption{Optical flow qualitative results on KITTI (train) using RAFT and our models. First row is for the baseline. Second and third rows are outputs of models trained with either DB or OA loss. Bottom row shows our proposed method with multiplicative combination. 
}
\vspace{-10pt}
\label{exp:qualitative_of}
\end{figure*}

\begin{table}[t]
\begin{center}
\caption{Optical flow results on Sintel (test) and KITTI (test) datasets. We finetune our model on Sintel, HD1K and KITTI.
}
\vspace{-4pt}
\label{tab:of_test}
\adjustbox{max width=0.38\textwidth}
{
\begin{tabular}{|c||c|c|c|}
\hline
 &  & RAFT & Ours \\
 \hline
\multirow{2}{*}{Sintel} & EPE-all & \textbf{1.609} & 1.685 \\
\multirow{2}{*}{(clean)} & EPE matched & 0.623 & \textbf{0.620} \\
 & EPE unmatched & \textbf{9.647} & 10.367 \\
 \hline
 \multirow{2}{*}{Sintel} & EPE-all & 2.855 & \textbf{2.837} \\
\multirow{2}{*}{(final)} & EPE matched & 1.405 & \textbf{1.300} \\
 & EPE unmatched & \textbf{14.680} & 15.367 \\
 \hline
  \multirow{3}{*}{KITTI} & Fl-bg & \textbf{4.74} & 4.77 \\
 & Fl-fg & 6.87 & \textbf{6.47} \\
 & Fl-all & 5.10 & \textbf{5.05} \\
\hline
\end{tabular}
}
\vspace{-20pt}
\end{center}
\end{table}

\section{Experiments}
\label{sec:experiment}

We evaluate our methods for both tasks of optical flow and stereo depth model on several datasets. We use RAFT~\cite{teed2020raft} and FlowFormer~\cite{huang2022flowformer} as our optical flow baseline architectures and RAFT-stereo~\cite{lipson2021raft} as our stereo depth baseline architecture. 

\subsection{Setup}
\textbf{Optical Flow Estimation: }
We follow the respective training protocols~\cite{teed2020raft, huang2022flowformer} and the hyperparameters from the RAFT and Florformer baselines. e.g. batch size, learning rate, number of training iterations, etc. We train our model on FlyingChairs (C)~\cite{dosovitskiy2015flownet} and FlyingThings3D (T)~\cite{mayer2016large} datasets and evaluate on Sintel (S) train~\cite{butler2012naturalistic} and KITTI (K) train~\cite{geiger2013vision, menze2015object} datasets. In addition, we finetune our model on Sintel (train), HD1K (H)~\cite{kondermann2016hci} and KITTI (train) datasets using C+T pre-trained model and evaluate on Sintel (test) and KITTI (test) datasets.  

\textbf{Stereo Depth Estimation: } 
We follow the baseline RAFT-stereo training protocol~\cite{lipson2021raft} with all the same hyperparameters. We train SceneFlow datasets (consists of FlyingThings3D, Monkaa~\cite{mayer2016large}, and Driving~\cite{mayer2016large}) and evaluate on ETH3D (train/test)~\cite{schops2017multi}, MiddlueBury~\cite{scharstein2014high}, Sintel (train), and KITTI (train) datasets. We also finetune our model on KITTI (train) dataset using sceneflow pre-trained model and evaluate on KITTI (test) dataset.

\subsection{Optical Flow Estimation}
Table~\ref{tab:of_training} shows the optical flow evaluation results on Sintel (train) and KITTI (train) datasets. Comparing with the RAFT baseline \footnote{For objectiveness, we train our baseline models in the same framework and report the results.}, models trained with either DB or OA loss demonstrate accuracy improvement, especially on the KITTI dataset. Among those four options, the particular combination of multiplication demonstrates the best performance. The differences between the standalone OA and the multiplicative combination are relatively minor. While OA is 0.01 better in EPE on Sintel (clean), it is 0.01 worse in EPE on Sintel (final). To investigate deeper, we apply additional metrics in Section \ref{sec:discussion} (Discussion) and provide more details. Overall, the multiplicative combination, among those four options of combination, demonstrates competitive accuracy against the baseline. Our combined DB and OA losses also outperforms the FlowFormer baseline.

\begin{table*}[ht]
\begin{center}
\caption{Stereo Depth estimation results on Eth3D, Middlebury, and KITTI (train) datasets. We train the model on SceneFlow. \textbf{Bold}/\underline{Underline}: Best and second best results. Errors are the percentage of pixels with EPE larger than the specific threshold. We follow the standard evaluation thresholds: 1px for ETH3D, 2px for Middlebury, and 3px for Sintel and KITTI. (* is tested by ourselves)
}
\vspace{-2mm}
\label{tab:dfs_training}
\adjustbox{max width=1.0\textwidth}
{
\begin{tabular}{|l|l||c|c|c|c|c|c|c|}
\hline
\multirow{2}{*}{Model} & \multirow{2}{*}{Method} & \multirow{2}{*}{ETH3D ($\downarrow$)} & \multicolumn{3}{|c|}{Middlebury ($\downarrow$)} & \multicolumn{2}{|c|}{Sintel (train) ($\downarrow$)} & KITTI ($\downarrow$) \\
\cline{4-8}
& & & F & H & Q & (clean) & (final) & (train) \\
\hline
\multirow{7}{*}{RAFT-Stereo} & Baseline & 3.28/3.26* & 18.33/18.43* & 12.59/11.56* & 9.36/10.00* & -/10.80* & -/12.45* & 5.74/6.12* \\
\cline{2-9}
& Difficulty Balancing (DB) & 2.65 & 17.05 & \textbf{10.48} & 8.50 & 10.51 & 12.57 & 4.52 \\
& Occlusion Avoiding (OA) & 3.38 & 17.66 & 11.07 & 9.20 & 10.54 & 12.57 & 5.63 \\
\cline{2-9}
& Combination (Sum) & \underline{2.61} & \underline{16.67} & 11.07 & 10.17 & 10.59 & 12.69 & 4.93 \\
& Combination (Multiplication) & 2.91 & 17.33 & \underline{10.54} & \textbf{7.88} & \underline{10.42} & \underline{12.30} & \textbf{4.25} \\
& Combination (Masking) & 2.69 & 18.49 & 12.45 & \underline{8.28} & 10.59 & 12.37 & 4.43 \\
& \cellcolor{lightgray}Combination (Mask-Sum) & \cellcolor{lightgray}\textbf{2.44} & \cellcolor{lightgray}\textbf{16.19} & \cellcolor{lightgray}12.64 & \cellcolor{lightgray}\textbf{7.88} & \cellcolor{lightgray}\textbf{10.24} & \cellcolor{lightgray}\textbf{12.00} &\cellcolor{lightgray}\underline{4.42} \\
\hline
\end{tabular}
}
\vspace{-3mm}
\end{center}
\end{table*}

\begin{table}[ht]
\begin{center}
\caption{Stereo Depth results on ETH3D and KITTI (test) dataset.
}
\vspace{-1mm}
\label{tab:dfs_test}
\adjustbox{max width=0.40\textwidth}
{
\begin{tabular}{|c||c|c|c|}
\hline
& & RAFT-Stereo  & Ours \\
\hline
\multirow{4}{*}{ETH3D} & bad 0.5 (\%) & 7.04 & \textbf{5.07} \\
& bad 1.0 (\%) & 2.44 & \textbf{1.67} \\
& bad 2.0(\%) & 0.44 & \textbf{0.39} \\
& AvgErr  & 0.18 & \textbf{0.15} \\
\hline
\multirow{3}{*}{KITTI} & Dl-all & 1.96 & \textbf{1.83} \\
& Dl-fg & 2.89 & \textbf{2.54} \\
& Dl-bg & 1.75 & \textbf{1.69} \\

\hline
\end{tabular}
}
\vspace{-20pt}
\end{center}
\end{table}

\begin{figure*}[t]
\begin{center}$
\centering
\begin{tabular}{cccc}
\vspace{-0.1cm}
\hspace{-0.3cm}
\rotatebox{90}{\quad \text{\small{\textbf{Baseline}}}} &
\hspace{-0.45cm} \includegraphics[width=5.75cm]{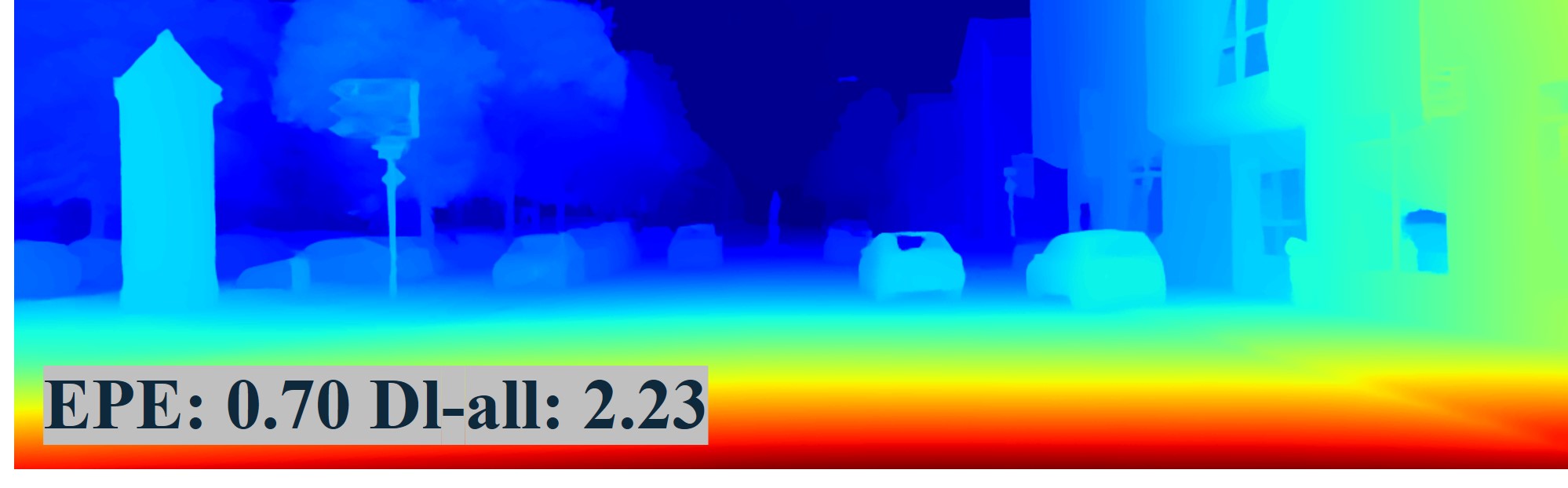}
& \hspace{-0.45cm} \includegraphics[width=5.75cm]{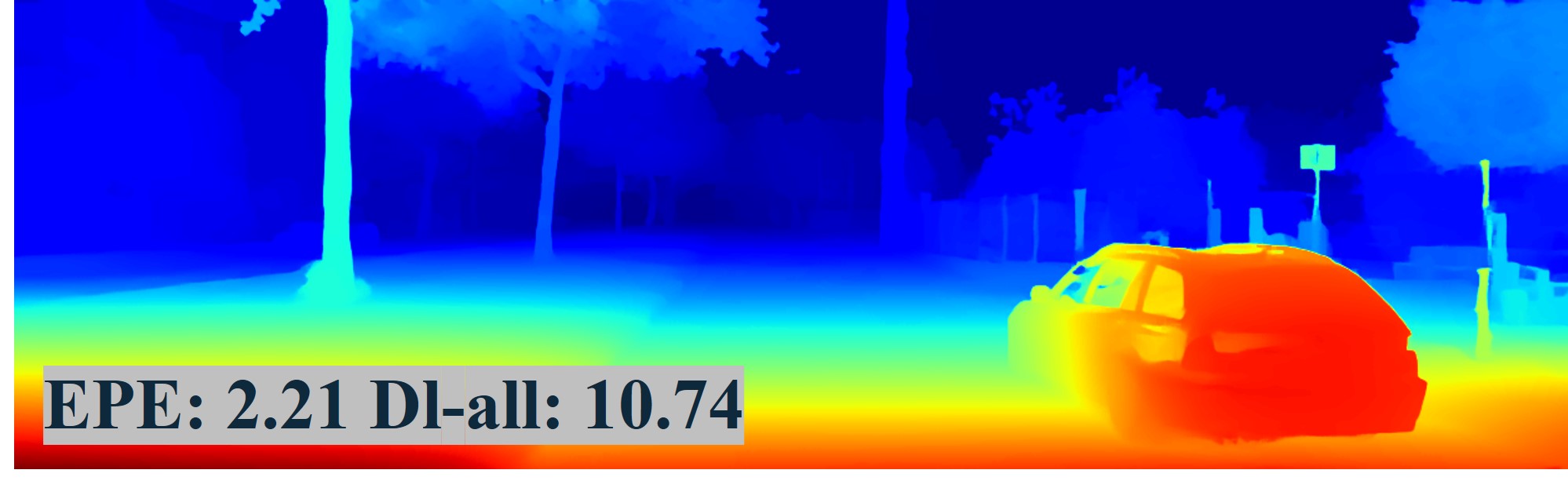}
& \hspace{-0.5cm} \includegraphics[width=5.75cm]{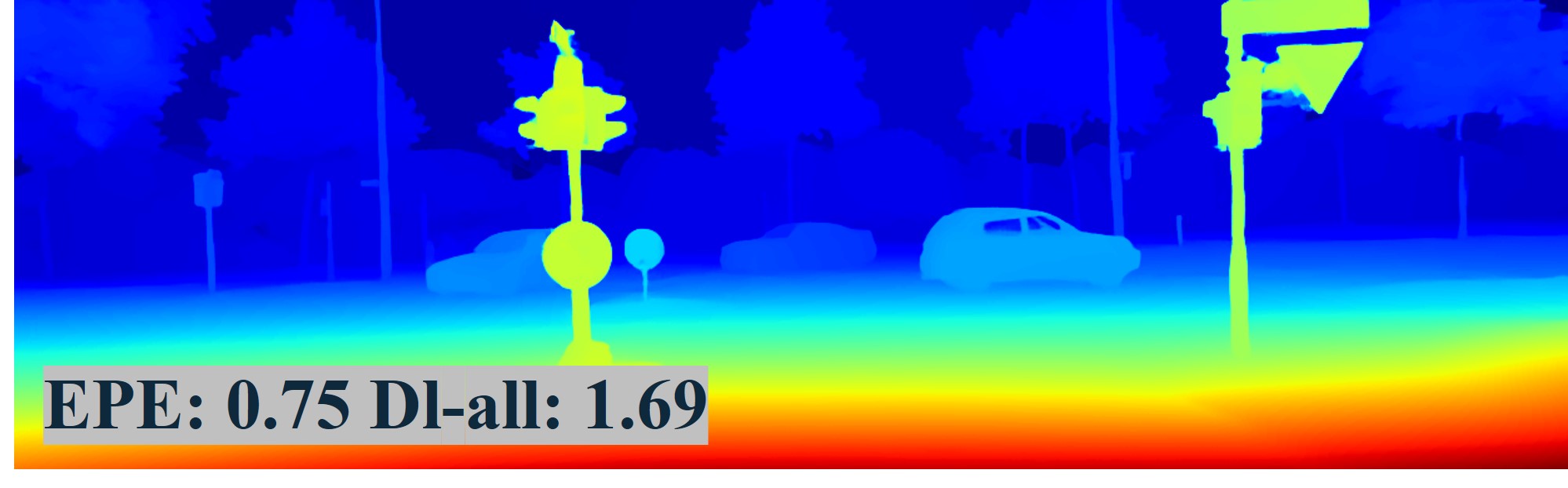} \\
\vspace{-0.1cm}
\hspace{-0.3cm} \rotatebox{90}{\qquad \ \text{\textbf{DB}}} &
\hspace{-0.45cm} \includegraphics[width=5.75cm]{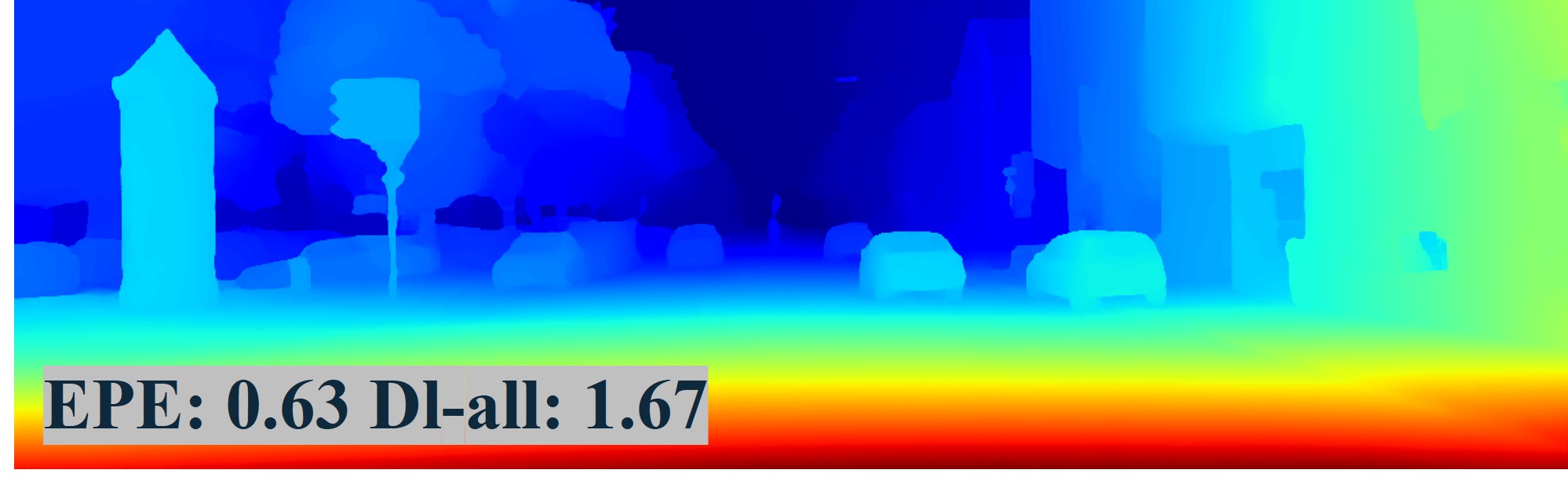}
& \hspace{-0.45cm} \includegraphics[width=5.75cm]{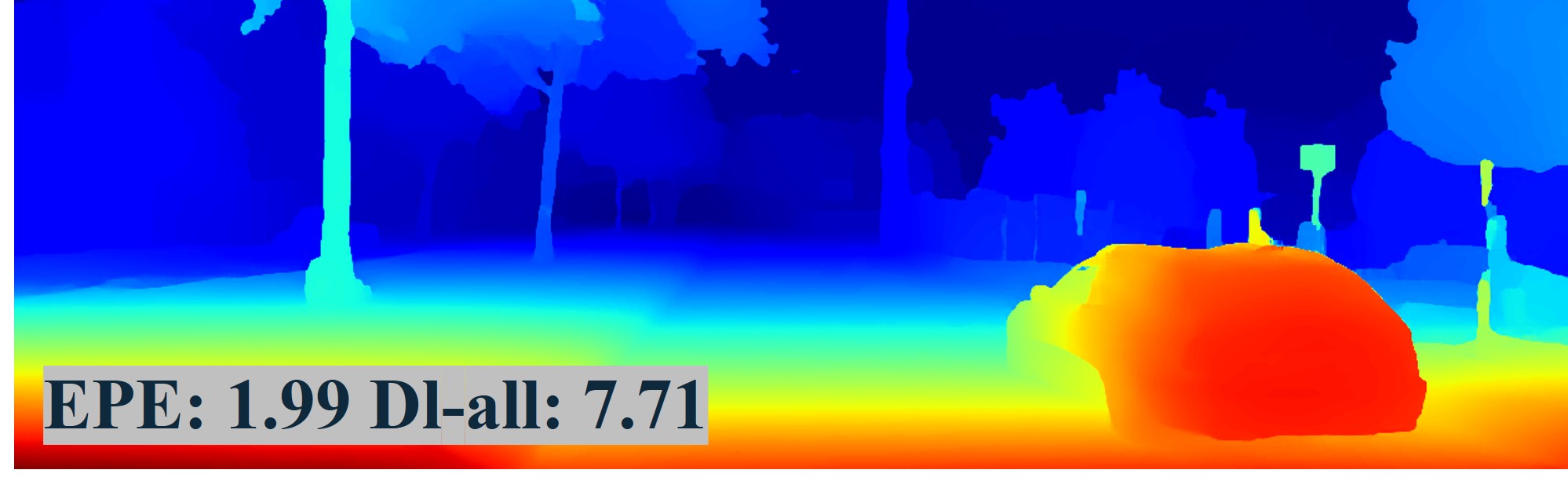}
& \hspace{-0.5cm} \includegraphics[width=5.75cm]{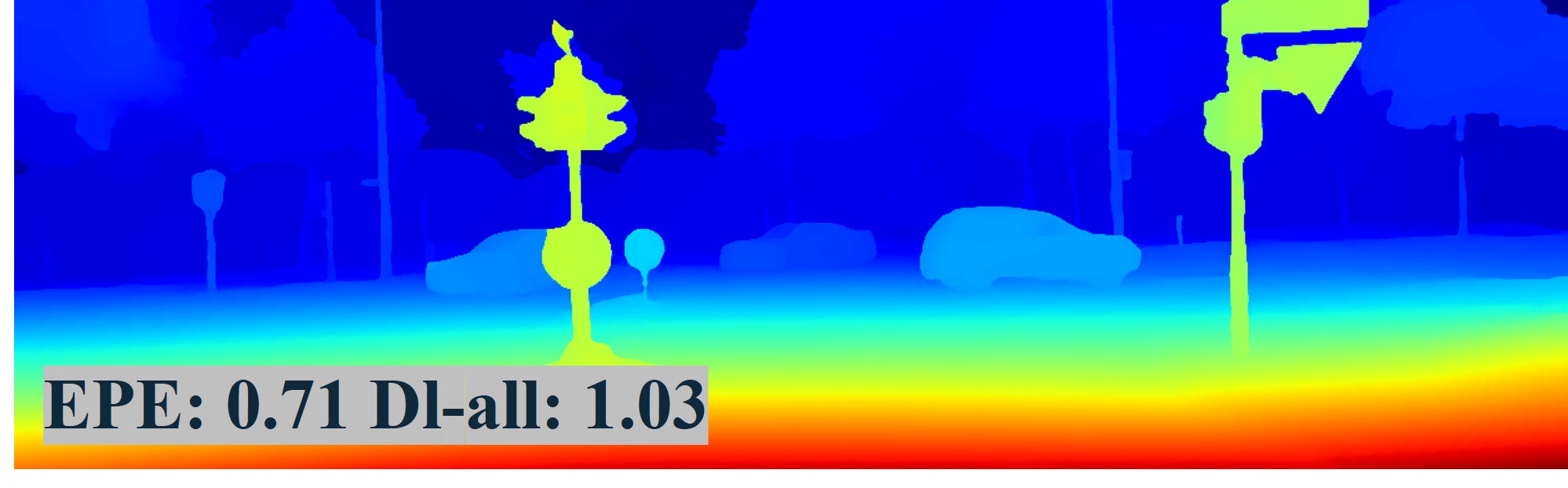} \\
\vspace{-0.1cm}
\hspace{-0.3cm}  \rotatebox{90}{\qquad \text{\textbf{OA}}} &
\hspace{-0.45cm} \includegraphics[width=5.75cm]{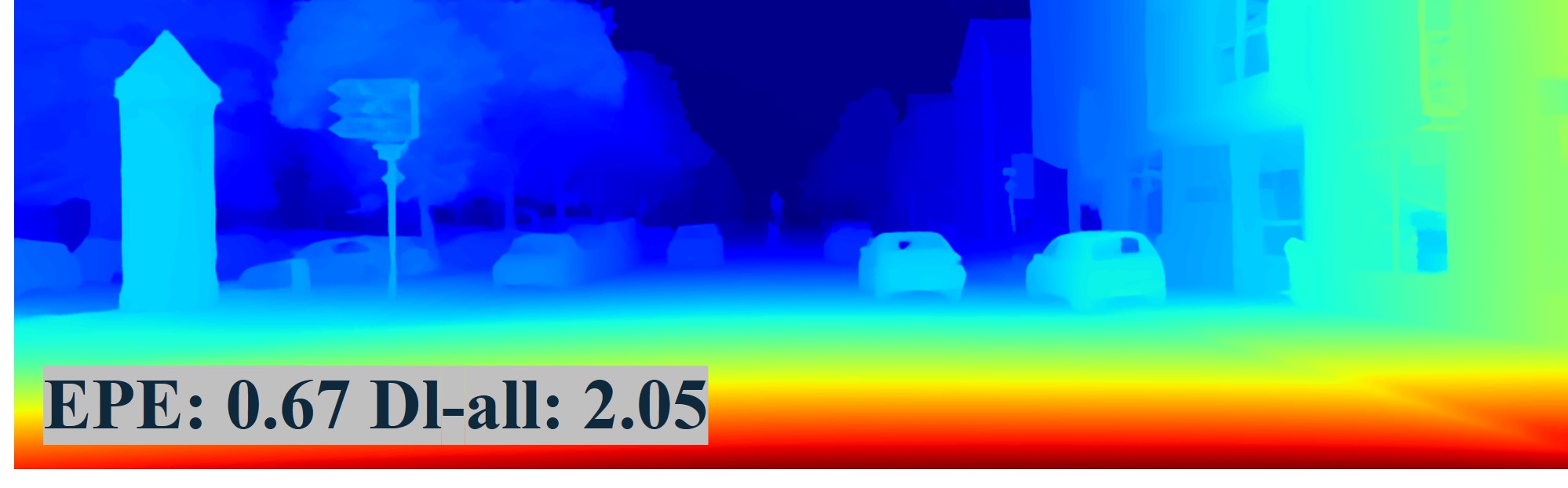}
& \hspace{-0.45cm} \includegraphics[width=5.75cm]{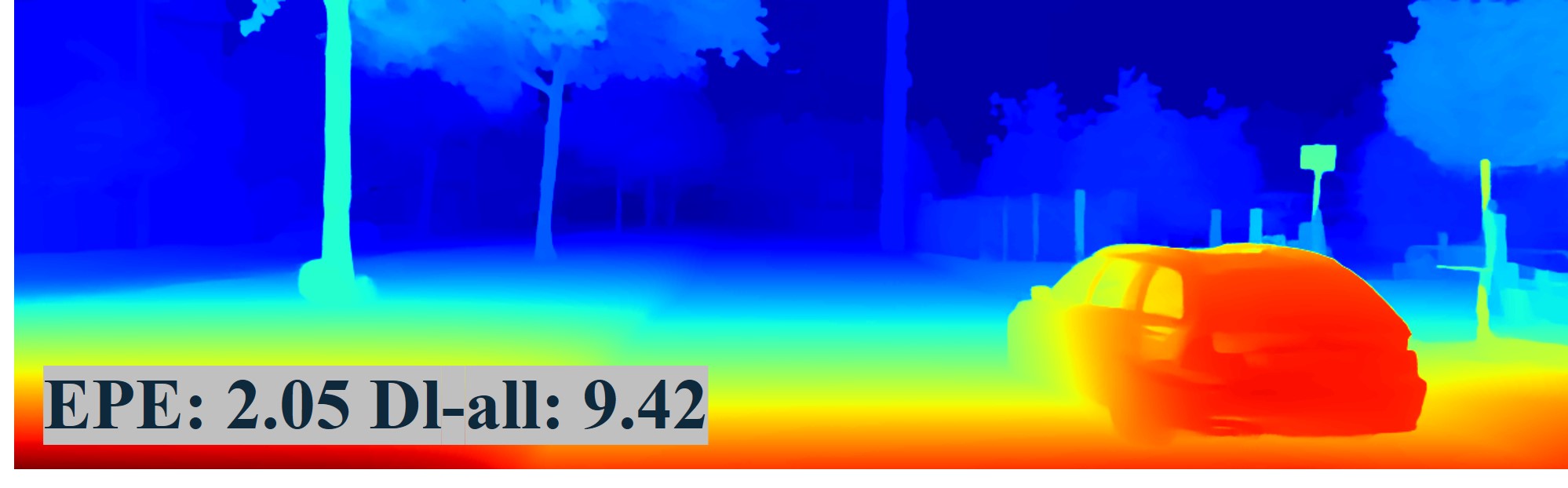}
& \hspace{-0.5cm} \includegraphics[width=5.75cm]{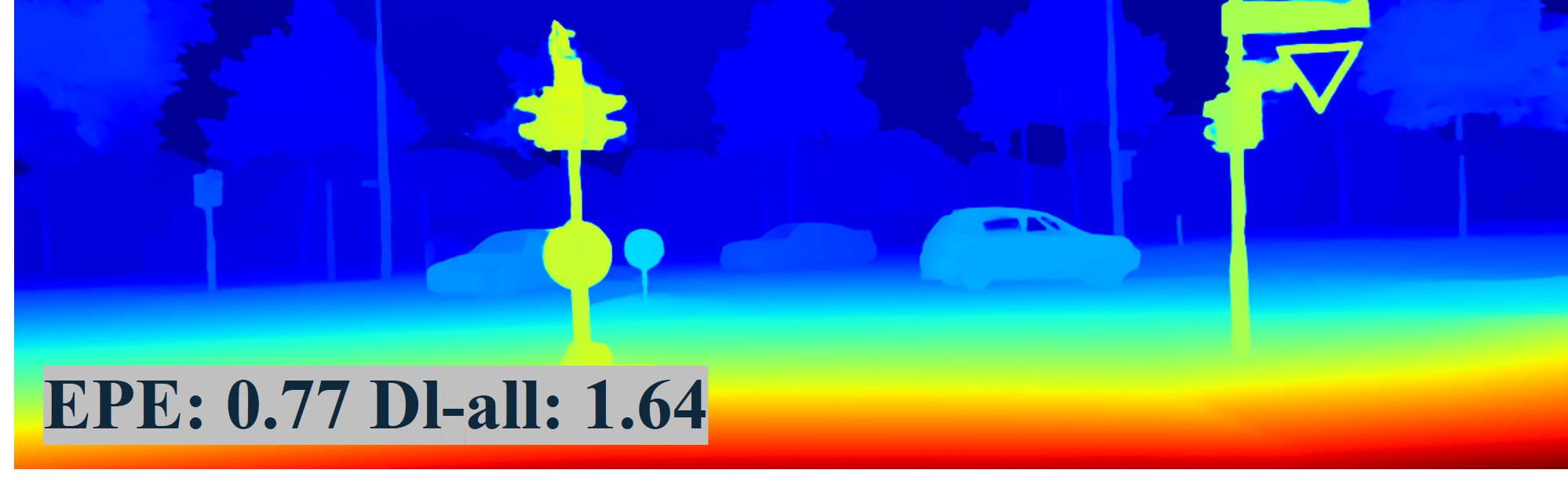} \\
\vspace{-0.1cm}
\hspace{-0.3cm} \rotatebox{90}{\quad \ \text{\textbf{Comb}}} &
\hspace{-0.45cm} \includegraphics[width=5.75cm]{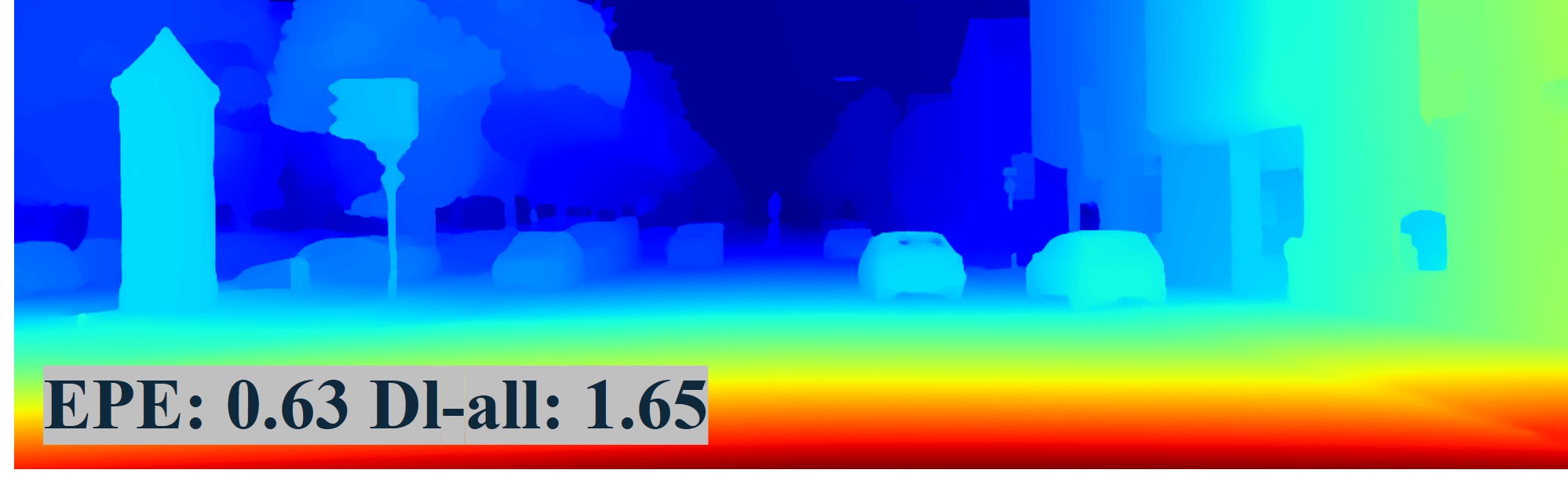}
& \hspace{-0.45cm} \includegraphics[width=5.75cm]{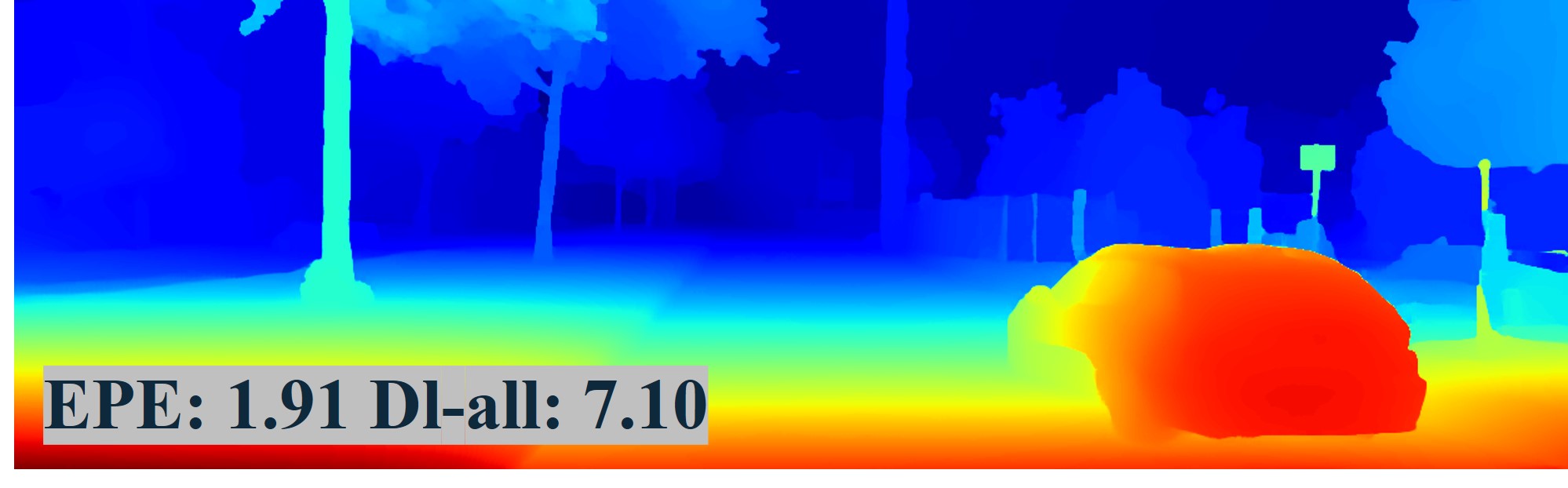}
& \hspace{-0.5cm} \includegraphics[width=5.75cm]{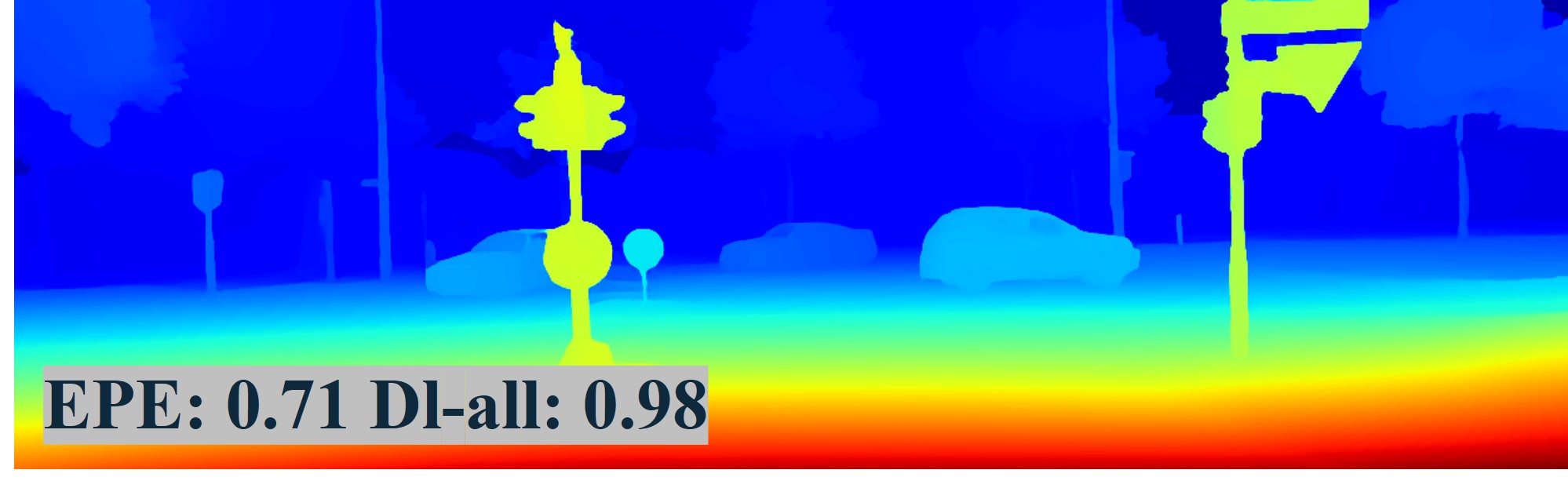} \\
\end{tabular}$
\end{center}
\vspace{-15pt}
\caption{Stereo Depth Qualitative results on KITTI (train) using RAFT-Stereo and our models. 
First row is the Baseline results. Second and third rows are outputs of models trained with each DB and OA loss. Bottom row shows our model trained with mask-sum combination.
}
\label{exp:qualitative_dfs}
\vspace{-5pt}
\end{figure*}

Table~\ref{tab:of_test} shows optical flow test results on Sintel (test) and KITTI (test) datasets. Due to the limitation in the number of tests, we finetune our model only with the multiplicative combination among those options. Our methods show significant improvement in the matching area, especially on Sintel (final), and Fl-foreground on the KITTI.

Figure~\ref{exp:qualitative_of} shows qualitative results on the KITTI dataset for the RAFT baseline and three models of our methods with DB, OA, and multiplicative combination on top of the RAFT baseline. DB shows better accuracy at object boundaries (tram in the left column, pillar of the right sign in the middle column), while OA shows better performance in the occluded areas (left vehicle in the right column). Our multiplicative combination (Comb) takes the advantages of both DB and OA and shows the best overall accuracy.

\subsection{Stereo Depth Estimation}
Table~\ref{tab:dfs_training} shows the stereo depth evaluation results on ETH3D, MiddleBury, Sintel and KITTI (train) datasets. Models trained with either DB or OA loss outperforms the baseline on the several benchmark datasets. Interestingly, the stereo depth model trained with the DB loss demonstrates particular accuracy gains over that with the OA loss, while in the case of optical flow model it is the opposite between the DB and OA losses. 
We hypothesize that, unlike optical flow estimation, stereo depth estimation involves the rectification operation over static objects and scenes, therefore the occlusion in the stereo image pair may be relatively straightforward. We also combine DB and OA losses, and the particular Masking-Sum combination shows the best performance overall for the stereo depth task.

\begin{table*}[t]
\begin{center}
\caption{Optical flow results on Sintel (train) datasets. We train the model on FlyingChairs (C) and FlyingThings3D (T). \textbf{Bold}/\underline{Underline}: Best and second best results. 1PX, 3PX, 5PX represent the percentage of pixels with EPE larger than 1 pixel, 3 pixel, and 5 pixel, respectively. $s_{0-10}$, $s_{10-40}$, and $s_{40+}$ represent the EPE with magnitude of ground truth less than 10, between 10 to 40, and larger than 40, respectively.
}
\vspace{-2mm}
\label{tab:analysis}
\adjustbox{max width=1.0\textwidth}
{
\begin{tabular}{|l||ccccccc|ccccccc|}
\hline
\multirow{3}{*}{Method} & \multicolumn{14}{|c|}{Sintel (train)}\\
\cline{2-15}
 & \multicolumn{7}{|c|}{clean}  & \multicolumn{7}{|c|}{final}\\
 & EPE & 1PX & 3PX & 5PX & $s_{0-10}$ & $s_{10-40}$ & $s_{40+}$ & EPE  & 1PX & 3PX & 5PX & $s_{0-10}$ & $s_{10-40}$ & $s_{40+}$ \\
\hline
RAFT (Baseline) & 1.43 & 9.84 & 4.41 & 3.17 & \underline{0.31} & \underline{1.52} & 9.21 & 2.69 & 14.72 & 8.09 & 6.19 & 0.51 & 2.96 & \underline{17.52} \\
\hline
Difficulty Balancing (DB) & 1.41 & 9.67 & 4.26 & 3.05 & 0.32 & 1.53 & 8.92 & 2.68 & 14.58 & 7.87 & 5.99 & 0.50 & 2.96 & \underline{17.52} \\
Occlusion Avoiding (OA) & \textbf{1.34} & 9.42 & 4.19 & 3.02 & \textbf{0.29} & \textbf{1.45} & 8.56 & 2.66 & 14.31 & 7.80 & 5.95 & \textbf{0.46} & 2.82 & 17.88\\
\hline
Combination (Sum) & 1.39 & 9.48 & 4.24 & 3.05 & \underline{0.31} & 1.54 & 8.77 & 2.69 & 14.15 & 7.69 & \underline{5.88} & \underline{0.48} & 2.94 & 17.85 \\
Combination (Multiplication) & \underline{1.35} & \textbf{9.06} & \textbf{4.13} & \textbf{2.99} & 0.32 & \underline{1.52} & \textbf{8.33} & \underline{2.65} & \textbf{13.88} & \textbf{7.54} & \textbf{5.72} & \underline{0.48} & \textbf{2.79} & 17.70 \\
Combination (Mask) & 1.38 & 9.25 & 4.17 & 3.03 & 0.33 & 1.58 & 8.41 & 2.71 & 14.25 & 7.78 & 5.94 & 0.49 & 2.97 & 17.88 \\
Combination (Mask-Sum) & 1.38 & 9.26 & 4.17 & 3.03 & 0.33 & 1.60 & 9.33 & \textbf{2.58} & \underline{14.04} & \underline{7.56} & \textbf{5.72} & \underline{0.48} & \underline{2.80} & \textbf{16.96} \\

\hline

\end{tabular}
}
\vspace{-8pt}
\end{center}
\end{table*}

\begin{figure*}[t]
\begin{center}$
\centering
\begin{tabular}{ccc}
\hspace{-0.3cm}
$M_{DB}$ & 1-$M_{DB}$ & $M_{OA}$\\
\hspace{-0.1cm}\includegraphics[width=5.7cm]{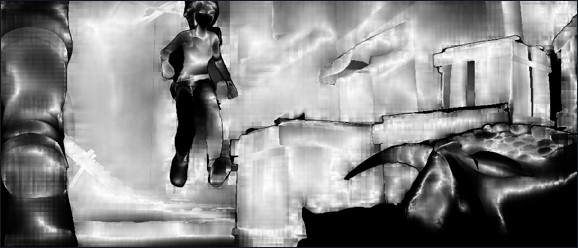} &
\hspace{-0.3cm}\includegraphics[width=5.7cm]{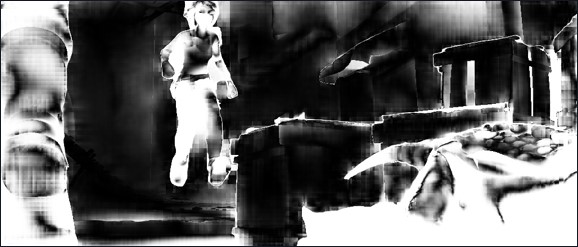}& 
\hspace{-0.3cm}\includegraphics[width=5.7cm]{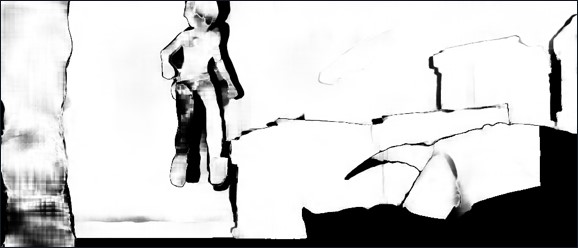} 
\end{tabular}$
\end{center}
\vspace{-12pt}
\caption{Confidence map results of $M_{DB}$ and $M_{OA}$. Error based confidence map $M_{DB}$ (left) is obtained by Eq.~\ref{ldb_confmap}, and $1-M_{DB}$ (middle) is used in the loss function. Forward backward consistency based confidence map $M_{OA}$ (right) is computed by \ref{oa_confmap}.}
\label{fig:conf_map}
\vspace{-3pt}
\end{figure*}

Table~\ref{tab:dfs_test} shows stereo depth test results on ETH3D and KITTI datasets. Since ETH3D dataset is relatively small, we evaluate our Sceneflow trained model on ETH3D test dataset. For KITTI evaluation, we finetune the RAFT-Stereo architecture with Mask-Sum combination loss and show improvements for all (foreground and background) on KITTI dataset. 

Figure~\ref{exp:qualitative_dfs} shows qualitative results on the KITTI dataset for the RAFT-Stereo baseline and three models with our methods of DB, OA, and Mask-Sum combination (Comb) on top of the baseline. Results with the DB loss show accurate performance on vehicle windows, while results with the OA loss did not show significant visual difference over the baseline. The Mask-sum combination (Comb) shows a minor improvement compared to DB results, demonstrating a 0.1 Dl-all difference on the KITTI (train) dataset.

\section{Discussion}
\label{sec:discussion}

\subsection{Analysis of Optical Flow Results using Additional Evaluation Metrics}
Table~\ref{tab:analysis} presents the optical flow results on the Sintel (train) dataset, including additional popular evaluation metrics. 1PX, 3PX, and 5PX denote the percentages of pixels where the End Point Error (EPE) exceeds 1, 3, and 5 pixels, respectively. Either of the DB and OA losses demonstrates reduction in the percentage of outliers in all cases, and the multiplicative combination in particular shows the lowest outlier rate in all cases. $s_{0-10}$, $s_{10-40}$, and $s_{40+}$ represent the EPE where the magnitude of the ground truth is less than 10, between 10 and 40, and greater than 40, respectively. The model trained with the DB loss show slight degradation at $s_{0-10}$ and $s_{10-40}$, while improving for $s_{40+}$. We hypothesize that small motions might be easier to train compared to large motions. The DB loss, which focuses on difficult regions, slightly underperforms for small displacements but it outperforms for larger displacements. The model trained with the OA loss performs best for small displacements ($s_{0-10}$), although not as robustly for larger displacements. The model trained with multiplicative combination in particular shows the best accuracy in some cases and perform well in most cases. Overall, at small displacements, there is a minor difference between the baseline and our proposed method, but for large displacements, a significant improvement of our proposal is observed, which helps reduce the overall End-Point Errors.

\begin{table*}[t]
\begin{center}
\caption{Ablation studies for $\alpha$ and $\beta$ of DB loss on optical flow task. We train the RAFT model on FlyingChairs (C) and FlyingThings3D (T) and evaluate on Sintel (train) and KITTI (train) datasets. (we show the Sintel (clean) EPE, Sintel (final) EPE, KITTI EPE, and KITTI Fl-all) \textbf{Bold}/\underline{Underline}: Best and second best results.
}
\vspace{-1mm}
\label{tab:ldb_ablation}
\adjustbox{max width=1.0\textwidth}
{
\begin{tabular}{|c|c||c|c|c|c|c|}
\hline
\multicolumn{2}{|c|}{RAFT}& \multicolumn{5}{|c|}{$\beta$} \\
\cline{3-7}
\multicolumn{2}{|c|}{with DB} & 0.25 & 0.5 & 1.0 & 2.0 & 5.0  \\
\hline
\multirow{4}{*}{$\alpha$} & 0.5 & 1.54 / 2.79 / 4.75 / 16.50 & 1.43 / 2.79 / 4.89 / 16.74 & 1.43 / 2.77 / 4.89 / 16.71 & 1.39 / 2.89 / 4.67 / 16.45 & 1.45 / \underline{2.70} / 5.13 / 16.93 \\
& 1.0 & 1.54 / 2.83 / 4.78 / 16.78 & 1.42 / 2.71 / 4.73 / 16.28  & \underline{1.38} / 2.72 / 4.70 / 16.12  & 1.43 / 2.80 / 4.71 / 16.34 & \textbf{1.36} / 2.81 / 4.84 / 16.64 \\
& 2.0 & 1.45 / \underline{2.70} / 4.70 / 16.57 & 1.41 / \cellcolor{lightgray} \textbf{2.68} / \underline{4.65} / \textbf{15.92} & 1.53 / 2.73 / \textbf{4.59} / 16.29 & 1.64 / 2.78 / 4.98 / 17.17 & 1.45 / 2.98 / 4.74 / \underline{16.08} \\
& 5.0 & 1.50 / 2.75 / 4.95 / 17.35 & 1.54 / 2.75 / 4.75 / 16.36 & 1.53 / 2.82 / 4.89 / 16.52 & 1.52 / 2.77 / 4.99 / 17.38 & 1.65 / 2.81 / 4.72 / 16.26 \\

\hline
\end{tabular}
}
\vspace{-12pt}
\end{center}
\end{table*}

\begin{table*}[t]
\begin{center}
\caption{Ablation study for $\alpha$ and $\beta$ of OA losses on optical flow task. We train the RAFT model on FlyingChairs (C) and FlyingThings3D (T) and evaluate on Sintel (train) and KITTI (train) datasets. (we show the Sintel (clean) EPE, Sintel (final) EPE, KITTI EPE, and KITTI Fl-all) \textbf{Bold}/\underline{Underline}: Best and second best results.
}
\vspace{-1mm}
\label{tab:oa_ablation}
\adjustbox{max width=0.9\textwidth}
{
\begin{tabular}{|c|c||c|c|c|c|}
\hline
\multicolumn{2}{|c|}{RAFT}& \multicolumn{4}{|c|}{$\beta$} \\
\cline{3-6}
\multicolumn{2}{|c|}{with OA} & 0.5 & 1.0 & 2.0 & 5.0  \\
\hline
\multirow{5}{*}{$\alpha$} & 0.5 & 1.44 / 2.74 / 4.82 / 16.96 & \underline{1.37} / 2.71 / 4.81 / 16.57  & 1.50 / 2.70 / 4.96 / 16.87  & 1.47 / 2.74 / 4.63 / 16.37 \\
& 1.0 & 1.42 / 2.67 / 4.58 / 15.81 & 1.38 / 2.68 / 4.77 / 15.98 & 1.38 / \underline{2.61} / 5.00 / 16.71 & 1.44 / 2.62 / 4.83 / 16.89 \\
& 2.0 & 1.49 / 2.78 / \textbf{4.34} / 16.18 & \cellcolor{lightgray}  \textbf{1.34} / 2.66 / \underline{4.44} / 15.77 & 1.41 / 2.66 / 4.44 / 15.85 & 1.63 / 2.71 / 5.02 / 17.24\\
& 5.0 & 1.44 / \textbf{2.55} / 4.59 / \underline{15.68} & 1.42 / 2.62 / 4.62 / \textbf{15.56} & 1.44 / 2.66 / 4.61 / 15.70 & 1.63 / 2.67 / 4.89 / 16.31 \\
& 10.0 & 1.50 / 2.71 / 5.04 / 16.96 & 1.44 / 2.67 / 4.82 / 16.01 & 1.46 / 2.70 / 4.89 / 16.82 & 1.47 / 2.62 / 4.77 / 16.57 \\

\hline
\end{tabular}
}
\vspace{-14pt}
\end{center}
\end{table*}

\subsection{Comparison of DB and OA Losses}
Figure~\ref{fig:conf_map} shows examples of DB and OA confidence maps. In DB loss, the actual weight map can be computed as $1-M_{DB}$ (middle of the Fig~\ref{fig:conf_map}). As shown in the middle weight map, the optical flow model predicts relatively accurate optical flows in the background, despite some errors in the foreground (e.g., objects or large displacements). The DB loss encourages the network to focus on difficult samples (e.g., large displacements or objects), improving on large motions as explained in the previous subsection. In contrast, the OA loss improves the overall accuracy by mitigating the effect of non-matching regions (right of the Fig~\ref{fig:conf_map}). The model trained with OA outperforms that with DB in all evaluation metrics, except for $s_{40+}$ on Sintel (final). The model trained with mask combination computes the DB loss only for non-occlusion regions. This result shows some improvements (Table~\ref{tab:analysis}), especially in the percentage of outliers, indicating that mitigating the occlusion effect in DB can enhance model performance.

\subsection{Combination of DB and OA Losses}
We apply four different combinations to model training, which may have different effects. In the summation combination, the occluded region can be compensated by each loss. For example, the occluded region in $1-M_{DB}$ (Bottom Right area) can have lower weights, while the corresponding region in $M_{OA}$ (Bottom Right area) may have higher weights. By combining these weights, their effects could be somewhat canceled out. Using a mask or multiplicative combination can mitigate such impact of the DB loss in occlusion areas, allowing the model to concentrate on difficult samples. In our experiments, multiplicative combination the best combination in optical flow, while Mask-Sum shows the best combination in stereo depth estimation.

\begin{table}[t]
\begin{center}
\caption{Ablation studies for $\alpha$ and $\beta$ of DB and OA losses on Stereo depth task. We train the RAFT-Stereo model on SceneFlow dataset and evaluate on ETH3D, Middlebury, and KITTI datasets.  
}
\vspace{-1mm}
\label{tab:dfs_training_ablation}
\adjustbox{max width=0.48\textwidth}
{
\begin{tabular}{|l||c|c|c|}
\hline
\multirow{2}{*}{Method} & \multirow{2}{*}{ETH3D ($\downarrow$)} & Middlebury ($\downarrow$) & KITTI ($\downarrow$) \\
& & F/H/Q & (train) \\
\hline
Baseline (Paper) & 3.28 & 18.33 / 12.59 / 9.36 & 5.74 \\
Baseline (Our) &  3.26 & 18.43 / 11.56 / 10.00 & 6.12 \\
\hline
DB ($\alpha$ = 2.0, $\beta$ = 0.5) & 2.66 & 17.47 / 10.54 / 8.83 & 4.57\\
DB ($\alpha$ = 2.0, $\beta$ = 1.0) & \textbf{2.65} & \textbf{17.05} / \textbf{10.48} / \textbf{8.50} & \textbf{4.52} \\
\hline
OA ($\alpha$ = 2.0, $\beta$ = 1.0) & \textbf{3.19} & 17.47 / 13.86 / 10.19 & 5.74 \\
OA ($\alpha$ = 1.0, $\beta$ = 1.0) & 3.38 & \textbf{17.66} / \textbf{11.07} / \textbf{9.20} & \textbf{5.63} \\
\hline
\end{tabular}
}
\vspace{-12pt}
\end{center}
\end{table}

\subsection{Ablation Study for $\alpha$ and $\beta$}
Table~\ref{tab:ldb_ablation} shows the optical flow results using the DB loss with various $\alpha$ and $\beta$. We adopt four different $\alpha$s (0.5, 1.0, 2.0, 5.0) and five different $\beta$s (0.25, 0.5, 1.0, 2.0, 5.0). In the table, if ($\alpha$, $\beta$) is (1.0, 1.0), it is the result of regression focal loss. Among these results, ($\alpha$ = 2.0, $\beta$ = 1.0) shows the best overall accuracy. The model with the best hyperparameters shows additional improvement over the RFL loss~\cite{lin2024sciflow}. We also find the best hyperparameter for stereo depth model training as shown in Table~\ref{tab:dfs_training_ablation}. We found that it shows the best performance at ($\alpha$ = 2.0 and $\beta$ = 1.0).

Table~\ref{tab:oa_ablation} shows the performance of optical flow using OA with various $\alpha$ and $\beta$ values. We apply five different $\alpha$s (0.5, 1.0, 2.0, 5.0, 10.0) and four different $\beta$s (0.5, 1.0, 2.0, 5.0) to RAFT-Stereo model. Among these, it shows good overall performance when ($\alpha$ and $\beta$) is (2.0 or 5.0 / 0.5 or 1.0). When $\beta$ is 0.5, it shows the best Sintel (final) or KITTI EPE score, but it underperforms the original RAFT on Sintel (clean). We choose ($\alpha$ =2.0 and $\beta$ = 1.0) because it demonstrates good performance, especially on Sintel (clean) dataset. We also apply different hyperparameter for stereo depth (TableTable~\ref{tab:dfs_training_ablation}). We found that $\alpha$ = 1.0 and $\beta$ = 1.0 shows better performance for OA loss.

\section{Conclusion}
\label{sec:conclusion}

In this paper, we have proposed novel confidence-based training methods effective for optical flow and stereo depth estimation. We have introduced the Difficulty Balancing (DB) loss, a unique approach that helps focus on challenging pixels through the introduction of tune-able hyperparameters, drawing inspiration from Regression Focal Loss. Furthermore, we have proposed the Occlusion Avoiding (OA) loss, which employs a stereo consistency-based confidence map to mitigate the challenge of non-matched regions during training. Recognizing the effectiveness of each loss, we have explored options of loss combinations to enhance the model learning. Our extensive experiments on standard optical flow and stereo depth benchmarks have not only demonstrated the effectiveness of individual losses, but also highlight the significant improvements achieved by their combination. This research, therefore, presents a significant advancement in the field of optical flow and stereo depth estimation.

\newpage
{
    \small
    \bibliographystyle{ieeenat_fullname}
    \bibliography{main}

\begin{thebibliography}{31}
\providecommand{\natexlab}[1]{#1}
\providecommand{\url}[1]{\texttt{#1}}
\expandafter\ifx\csname urlstyle\endcsname\relax
  \providecommand{\doi}[1]{doi: #1}\else
  \providecommand{\doi}{doi: \begingroup \urlstyle{rm}\Url}\fi

\bibitem[Butler et~al.(2012)Butler, Wulff, Stanley, and Black]{butler2012naturalistic}
Daniel~J Butler, Jonas Wulff, Garrett~B Stanley, and Michael~J Black.
\newblock A naturalistic open source movie for optical flow evaluation.
\newblock In \emph{Proceedings of the European Conference on Computer Vision}, pages 611--625. Springer, 2012.

\bibitem[Cai et~al.(2019)Cai, Neher, Vats, Clausi, and Zelek]{cai2019temporal}
Zixi Cai, Helmut Neher, Kanav Vats, David~A Clausi, and John Zelek.
\newblock Temporal hockey action recognition via pose and optical flows.
\newblock In \emph{Proceedings of the IEEE/CVF Conference on Computer Vision and Pattern Recognition Workshops}, pages 0--0, 2019.

\bibitem[Dosovitskiy et~al.(2015)Dosovitskiy, Fischer, Ilg, Hausser, Hazirbas, Golkov, Van Der~Smagt, Cremers, and Brox]{dosovitskiy2015flownet}
Alexey Dosovitskiy, Philipp Fischer, Eddy Ilg, Philip Hausser, Caner Hazirbas, Vladimir Golkov, Patrick Van Der~Smagt, Daniel Cremers, and Thomas Brox.
\newblock Flownet: Learning optical flow with convolutional networks.
\newblock In \emph{Proceedings of the IEEE/CVF International Conference on Computer Vision}, pages 2758--2766, 2015.

\bibitem[Geiger et~al.(2013)Geiger, Lenz, Stiller, and Urtasun]{geiger2013vision}
Andreas Geiger, Philip Lenz, Christoph Stiller, and Raquel Urtasun.
\newblock Vision meets robotics: The kitti dataset.
\newblock \emph{The International Journal of Robotics Research}, 32\penalty0 (11):\penalty0 1231--1237, 2013.

\bibitem[Huang et~al.(2022)Huang, Shi, Zhang, Wang, Cheung, Qin, Dai, and Li]{huang2022flowformer}
Zhaoyang Huang, Xiaoyu Shi, Chao Zhang, Qiang Wang, Ka~Chun Cheung, Hongwei Qin, Jifeng Dai, and Hongsheng Li.
\newblock Flowformer: A transformer architecture for optical flow.
\newblock In \emph{Proceedings of the European Conference on Computer Vision}, 2022.

\bibitem[Hui and Loy(2020)]{hui2020liteflownet3}
Tak-Wai Hui and Chen~Change Loy.
\newblock Liteflownet3: Resolving correspondence ambiguity for more accurate optical flow estimation.
\newblock In \emph{Computer Vision--ECCV 2020: 16th European Conference, Glasgow, UK, August 23--28, 2020, Proceedings, Part XX 16}, pages 169--184. Springer, 2020.

\bibitem[Jeong et~al.(2023)Jeong, Cai, Garrepalli, and Porikli]{jeong2023distractflow}
Jisoo Jeong, Hong Cai, Risheek Garrepalli, and Fatih Porikli.
\newblock Distractflow: Improving optical flow estimation via realistic distractions and pseudo-labeling.
\newblock In \emph{Proceedings of the IEEE/CVF Conference on Computer Vision and Pattern Recognition}, pages 13691--13700, 2023.

\bibitem[Jeong et~al.(2024)Jeong, Cai, Garrepalli, Lin, Hayat, and Porikli]{jeong2024ocai}
Jisoo Jeong, Hong Cai, Risheek Garrepalli, Jamie~Menjay Lin, Munawar Hayat, and Fatih Porikli.
\newblock Ocai: Improving optical flow estimation by occlusion and consistency aware interpolation.
\newblock In \emph{Proceedings of the IEEE/CVF Conference on Computer Vision and Pattern Recognition}, pages 19352--19362, 2024.

\bibitem[Jiang et~al.(2021)Jiang, Campbell, Lu, Li, and Hartley]{jiang2021learning}
Shihao Jiang, Dylan Campbell, Yao Lu, Hongdong Li, and Richard Hartley.
\newblock Learning to estimate hidden motions with global motion aggregation.
\newblock In \emph{Proceedings of the IEEE/CVF International Conference on Computer Vision}, pages 9772--9781, 2021.

\bibitem[Jonschkowski et~al.(2020)Jonschkowski, Stone, Barron, Gordon, Konolige, and Angelova]{jonschkowski2020matters}
Rico Jonschkowski, Austin Stone, Jonathan~T Barron, Ariel Gordon, Kurt Konolige, and Anelia Angelova.
\newblock What matters in unsupervised optical flow.
\newblock In \emph{Proceedings of the European Conference on Computer Vision}, pages 557--572. Springer, 2020.

\bibitem[Kale et~al.(2015)Kale, Pawar, and Dhulekar]{kale2015moving}
Kiran Kale, Sushant Pawar, and Pravin Dhulekar.
\newblock Moving object tracking using optical flow and motion vector estimation.
\newblock In \emph{2015 4th international conference on reliability, infocom technologies and optimization (ICRITO)(trends and future directions)}, pages 1--6. IEEE, 2015.

\bibitem[Karaev et~al.(2023)Karaev, Rocco, Graham, Neverova, Vedaldi, and Rupprecht]{karaev2023dynamicstereo}
Nikita Karaev, Ignacio Rocco, Benjamin Graham, Natalia Neverova, Andrea Vedaldi, and Christian Rupprecht.
\newblock Dynamicstereo: Consistent dynamic depth from stereo videos.
\newblock In \emph{Proceedings of the IEEE/CVF Conference on Computer Vision and Pattern Recognition}, pages 13229--13239, 2023.

\bibitem[Kondermann et~al.(2016)Kondermann, Nair, Honauer, Krispin, Andrulis, Brock, Gussefeld, Rahimimoghaddam, Hofmann, Brenner, et~al.]{kondermann2016hci}
Daniel Kondermann, Rahul Nair, Katrin Honauer, Karsten Krispin, Jonas Andrulis, Alexander Brock, Burkhard Gussefeld, Mohsen Rahimimoghaddam, Sabine Hofmann, Claus Brenner, et~al.
\newblock The hci benchmark suite: Stereo and flow ground truth with uncertainties for urban autonomous driving.
\newblock In \emph{Proceedings of the IEEE/CVF Conference on Computer Vision and Pattern Recognition Workshops}, pages 19--28, 2016.

\bibitem[Kong et~al.(2022)Kong, Jiang, Luo, Chu, Huang, Tai, Wang, and Yang]{kong2022ifrnet}
Lingtong Kong, Boyuan Jiang, Donghao Luo, Wenqing Chu, Xiaoming Huang, Ying Tai, Chengjie Wang, and Jie Yang.
\newblock Ifrnet: Intermediate feature refine network for efficient frame interpolation.
\newblock In \emph{Proceedings of the IEEE/CVF Conference on Computer Vision and Pattern Recognition}, pages 1969--1978, 2022.

\bibitem[Lee et~al.(2018)Lee, Lee, Son, Park, and Kwak]{lee2018motion}
Myunggi Lee, Seungeui Lee, Sungjoon Son, Gyutae Park, and Nojun Kwak.
\newblock Motion feature network: Fixed motion filter for action recognition.
\newblock In \emph{Proceedings of the European Conference on Computer Vision}, pages 387--403, 2018.

\bibitem[Li et~al.(2022)Li, Wang, Xiong, Cai, Yan, Yang, Liu, Fan, and Liu]{li2022practical}
Jiankun Li, Peisen Wang, Pengfei Xiong, Tao Cai, Ziwei Yan, Lei Yang, Jiangyu Liu, Haoqiang Fan, and Shuaicheng Liu.
\newblock Practical stereo matching via cascaded recurrent network with adaptive correlation.
\newblock In \emph{Proceedings of the IEEE/CVF conference on computer vision and pattern recognition}, pages 16263--16272, 2022.

\bibitem[Li et~al.(2023)Li, Zhu, Han, Hou, Guo, and Cheng]{li2023amt}
Zhen Li, Zuo-Liang Zhu, Ling-Hao Han, Qibin Hou, Chun-Le Guo, and Ming-Ming Cheng.
\newblock Amt: All-pairs multi-field transforms for efficient frame interpolation.
\newblock In \emph{Proceedings of the IEEE/CVF Conference on Computer Vision and Pattern Recognition}, pages 9801--9810, 2023.

\bibitem[Lin et~al.(2024)Lin, Jeong, Cai, Garrepalli, Wang, and Porikli]{lin2024sciflow}
Jamie~Menjay Lin, Jisoo Jeong, Hong Cai, Risheek Garrepalli, Kai Wang, and Fatih Porikli.
\newblock Sciflow: Empowering lightweight optical flow models with self-cleaning iterations.
\newblock In \emph{Proceedings of the IEEE/CVF Conference on Computer Vision and Pattern Recognition}, pages 2162--2171, 2024.

\bibitem[Lipson et~al.(2021)Lipson, Teed, and Deng]{lipson2021raft}
Lahav Lipson, Zachary Teed, and Jia Deng.
\newblock Raft-stereo: Multilevel recurrent field transforms for stereo matching.
\newblock In \emph{2021 International Conference on 3D Vision (3DV)}, pages 218--227. IEEE, 2021.

\bibitem[Lu et~al.(2019)Lu, Ouyang, Xu, Zhang, Cai, and Gao]{lu2019dvc}
Guo Lu, Wanli Ouyang, Dong Xu, Xiaoyun Zhang, Chunlei Cai, and Zhiyong Gao.
\newblock Dvc: An end-to-end deep video compression framework.
\newblock In \emph{Proceedings of the IEEE/CVF Conference on Computer Vision and Pattern Recognition}, pages 11006--11015, 2019.

\bibitem[Mayer et~al.(2016)Mayer, Ilg, Hausser, Fischer, Cremers, Dosovitskiy, and Brox]{mayer2016large}
Nikolaus Mayer, Eddy Ilg, Philip Hausser, Philipp Fischer, Daniel Cremers, Alexey Dosovitskiy, and Thomas Brox.
\newblock A large dataset to train convolutional networks for disparity, optical flow, and scene flow estimation.
\newblock In \emph{Proceedings of the IEEE/CVF Conference on Computer Vision and Pattern Recognition}, pages 4040--4048, 2016.

\bibitem[Meister et~al.(2018)Meister, Hur, and Roth]{meister2018unflow}
Simon Meister, Junhwa Hur, and Stefan Roth.
\newblock Unflow: Unsupervised learning of optical flow with a bidirectional census loss.
\newblock In \emph{Proceedings of the AAAI conference on artificial intelligence}, 2018.

\bibitem[Menze and Geiger(2015)]{menze2015object}
Moritz Menze and Andreas Geiger.
\newblock Object scene flow for autonomous vehicles.
\newblock In \emph{Proceedings of the IEEE/CVF Conference on Computer Vision and Pattern Recognition}, pages 3061--3070, 2015.

\bibitem[Scharstein et~al.(2014)Scharstein, Hirschm{\"u}ller, Kitajima, Krathwohl, Ne{\v{s}}i{\'c}, Wang, and Westling]{scharstein2014high}
Daniel Scharstein, Heiko Hirschm{\"u}ller, York Kitajima, Greg Krathwohl, Nera Ne{\v{s}}i{\'c}, Xi Wang, and Porter Westling.
\newblock High-resolution stereo datasets with subpixel-accurate ground truth.
\newblock In \emph{Pattern Recognition: 36th German Conference, GCPR 2014, M{\"u}nster, Germany, September 2-5, 2014, Proceedings 36}, pages 31--42. Springer, 2014.

\bibitem[Schops et~al.(2017)Schops, Schonberger, Galliani, Sattler, Schindler, Pollefeys, and Geiger]{schops2017multi}
Thomas Schops, Johannes~L Schonberger, Silvano Galliani, Torsten Sattler, Konrad Schindler, Marc Pollefeys, and Andreas Geiger.
\newblock A multi-view stereo benchmark with high-resolution images and multi-camera videos.
\newblock In \emph{Proceedings of the IEEE conference on computer vision and pattern recognition}, pages 3260--3269, 2017.

\bibitem[Sohn et~al.(2020)Sohn, Berthelot, Carlini, Zhang, Zhang, Raffel, Cubuk, Kurakin, and Li]{sohn2020fixmatch}
Kihyuk Sohn, David Berthelot, Nicholas Carlini, Zizhao Zhang, Han Zhang, Colin~A Raffel, Ekin~Dogus Cubuk, Alexey Kurakin, and Chun-Liang Li.
\newblock Fixmatch: Simplifying semi-supervised learning with consistency and confidence.
\newblock \emph{Advances in neural information processing systems}, 33:\penalty0 596--608, 2020.

\bibitem[Stone et~al.(2021)Stone, Maurer, Ayvaci, Angelova, and Jonschkowski]{stone2021smurf}
Austin Stone, Daniel Maurer, Alper Ayvaci, Anelia Angelova, and Rico Jonschkowski.
\newblock Smurf: Self-teaching multi-frame unsupervised raft with full-image warping.
\newblock In \emph{Proceedings of the IEEE/CVF Conference on Computer Vision and Pattern Recognition}, pages 3887--3896, 2021.

\bibitem[Teed and Deng(2020)]{teed2020raft}
Zachary Teed and Jia Deng.
\newblock Raft: Recurrent all-pairs field transforms for optical flow.
\newblock In \emph{Proceedings of the European Conference on Computer Vision}, pages 402--419. Springer, 2020.

\bibitem[Wu et~al.(2018)Wu, Singhal, and Krahenbuhl]{wu2018video}
Chao-Yuan Wu, Nayan Singhal, and Philipp Krahenbuhl.
\newblock Video compression through image interpolation.
\newblock In \emph{Proceedings of the European Conference on Computer Vision}, pages 416--431, 2018.

\bibitem[Zhang et~al.(2021)Zhang, Woodford, Prisacariu, and Torr]{zhang2021separable}
Feihu Zhang, Oliver~J Woodford, Victor~Adrian Prisacariu, and Philip~HS Torr.
\newblock Separable flow: Learning motion cost volumes for optical flow estimation.
\newblock In \emph{Proceedings of the IEEE/CVF International Conference on Computer Vision}, pages 10807--10817, 2021.

\bibitem[Zhao et~al.(2020)Zhao, Sheng, Dong, Chang, Xu, et~al.]{zhao2020maskflownet}
Shengyu Zhao, Yilun Sheng, Yue Dong, Eric~I Chang, Yan Xu, et~al.
\newblock Maskflownet: Asymmetric feature matching with learnable occlusion mask.
\newblock In \emph{Proceedings of the IEEE/CVF Conference on Computer Vision and Pattern Recognition}, pages 6278--6287, 2020.

\end{thebibliography}
}


\end{document}